%% file: main.tex
\documentclass[]{x2lab}
\usepackage{csquotes}
\usepackage{afterpage}
\usepackage{amsmath,amssymb,amsfonts}
\usepackage{mathtools}
\usepackage{amsthm}
\usepackage{xspace}
\usepackage{colortbl}
\usepackage{enumitem}
\usepackage{wrapfig}
\usepackage{nicefrac}
\usepackage{makecell}
\usepackage{pifont}
\usepackage{svg}
\usepackage{array}
\usepackage{tabularx}
\usepackage{longtable}
\usepackage{xltabular}
\usepackage{float}
\usepackage{placeins}
\usepackage[utf8]{inputenc}
\usepackage{times}

\graphicspath{{figures/}{figs/}}

\definecolor{fbApp}{HTML}{c8e7fa}
\definecolor{fbPurple3}{HTML}{f0ebf5}
\definecolor{citecolor}{HTML}{0071BC}
\definecolor{linkcolor}{HTML}{ED1C24}

\newcommand{\pimodel}{\ensuremath{\pi_{0.5}}}
\newcommand{\pizero}{\ensuremath{\pi_{0}}}
\newcommand{\walloss}{Wall-OSS-0.5}

\titleformat*{\paragraph}{\rmfamily\bfseries}

\title{\begin{center}
Wall-OSS-0.5 Technical Report\\[-0.5cm]
{\large\mdseries\itshape Pretrain Once, Act Anywhere}
{\author{X Square Robot Team}}
\end{center}}

\abstract{Large-scale Vision-Language-Action (VLA) pretraining is increasingly adopted as the foundation for robot policies, yet the evidence for pretrained VLAs is almost invariably reported after task-specific fine-tuning.
This leaves a foundational question unanswered: does VLA pretraining itself yield executable robot behavior, or does it merely furnish a better initialization for downstream policy learning? 
We present \textbf{\walloss{}}, an open-source 4B VLA built upon a 3B VLM backbone augmented with action-generation components, designed so that pretrained robotic capability is directly measurable on physical hardware.
The model is pretrained across more than 20 embodiments, processing over one million robot trajectories per epoch alongside a grounded multimodal corpus.
We adopt a \emph{gradient-bridged co-training} recipe in which three objectives play distinct and complementary roles: discrete action prediction routes strong VLM-native gradients into the backbone, multimodal prediction preserves grounded vision-language understanding, and continuous flow matching serves as the deployment-time action interface. 
\emph{Before task-specific fine-tuning}, the pretrained checkpoint achieves non-trivial zero-shot real-robot behavior, completing several tasks, including a held-out deformable manipulation task, at high task progress on a 17-task suite. 
\emph{After fine-tuning}, the same checkpoint serves as a stronger adaptation prior, reaching 60.5\% average task progress on 15 real-robot tasks and outperforming \pimodel{}~\citep{pi05} by \textbf{17.5\%}.
Multimodal evaluations further confirm that action training does not erode grounded vision-language competence: the model preserves broad vision-language ability while strengthening embodied grounding. 
Together, these results reposition VLA pretraining from an initialization strategy to a directly testable, already useful source of robot capability.}

\metadata[Code]{\url{https://github.com/X-Square-Robot/wall-x}}
\metadata[Project Page]{\url{https://x2robot.com/en/oss}}

\begin{document}
\maketitle

\input{sec/1_intro}
\input{sec/2_methods}

\input{sec/3_data}
\input{sec/4_experiments}
\input{sec/5_ablation}
\input{sec/6_related}

\input{sec/7_discussion}
\input{sec/contributors}

\clearpage
\newpage
\bibliographystyle{unsrtnat}
\bibliography{main}

\clearpage
\appendix
\FloatBarrier
\input{sec/appendix}

\end{document}

%% file: sec/1_intro.tex
\section{Introduction}\label{sec:intro}

\begin{figure}[!t]
    \centering
    \includegraphics[width=\linewidth]{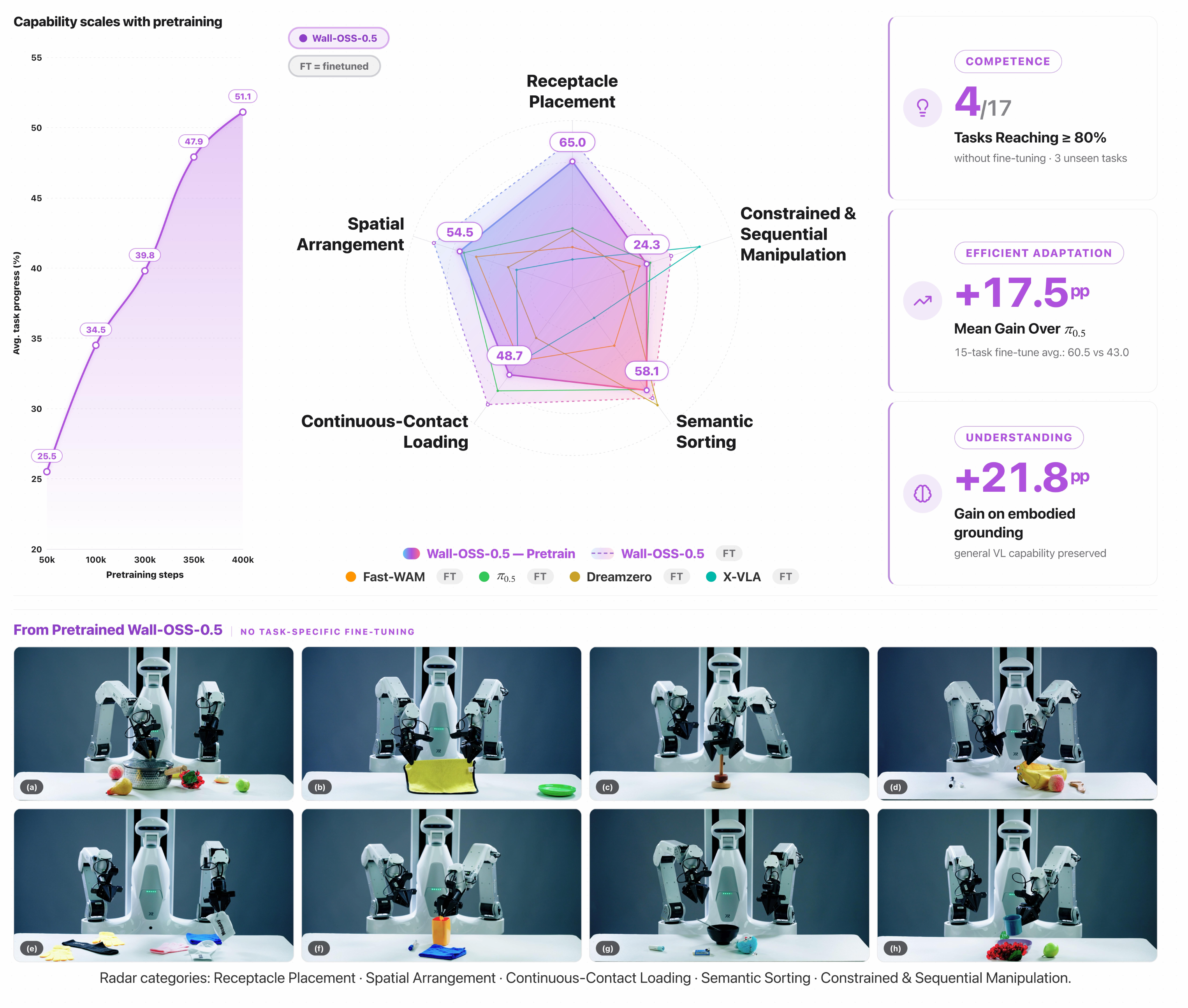}
    \caption{\textbf{\walloss{} capability overview.} The figure summarizes \walloss{} along three axes: pretrained model's real-robot behavior, downstream adaptation, and embodied multimodal understanding. Panels~(a)--(h) show examples of the evaluated real-robot tasks and their instructions: \textbf{(a)} open the pot lid; \textbf{(b)} cover the blocks; \textbf{(c)} stack the ring; \textbf{(d)} put the pen into the bag; \textbf{(e)} pair the socks; \textbf{(f)} insert the screwdriver; \textbf{(g)} put the spoon into the bowl; \textbf{(h)} put the cup onto the plate. Notably, when deployed directly without any task-specific fine-tuning, the pretrained model already executes many of these tasks at high task progress. \textit{Throughout, pp denotes percentage points.}}
    \label{fig:overview}
\end{figure}

Foundation models become credible when pretraining yields capabilities that are observable prior to task-specific adaptation.
In language and vision, this standard is by now routine: a pretrained model is expected to answer questions, follow instructions, and transfer to novel tasks.
In robotics, however, the evidence remains less direct.
Vision-Language-Action (VLA) models~\citep{zhai2025igniting,rt2,openvla,pi0,pi05,gemini_robotics,helix,xvla,team2026gigabrain,lingbot_vla,galaxea_openworld,wen2025tinyvla} increasingly inherit perception and reasoning from pretrained VLMs~\citep{Qwen2.5-VL,beyer2024paligemma}, yet their strongest results are typically reported only after downstream fine-tuning.
This raises a concrete operational question: \emph{does VLA pretraining alone produce an executable policy under real-robot evaluation?}

We introduce \textbf{\walloss{}}, an open-source 4B VLA built upon a 3B VLM backbone augmented with action-generation components, and organized around precisely this deployment-oriented criterion.
The pretrained checkpoint is evaluated directly as a real-robot policy, rather than merely as an initialization.
This imposes three requirements on the model: it must execute useful manipulation skills out of the box, retain sufficient VLM-derived vision-language competence to remain instruction-grounded, and furnish priors that render downstream adaptation more sample-efficient.
We refer to this objective as \emph{deployment-oriented VLA pretraining}.

The central empirical finding is that pretraining alone already produces measurable robot behavior, as shown in \Cref{fig:overview}.
On a 17-task real-robot zero-shot suite spanning semantic, rigid-object, deformable-object, fine-grained, and long-horizon manipulation, the pretrained checkpoint completes several tasks with high task progress in the absence of any task-specific fine-tuning: Block Sorting (100\%), Fruit Sorting (96\%), Ring Stacking (86\%), and the held-out deformable task Rope Tightening (82\%).
Held-out tasks track the learning curves of seen tasks throughout pretraining, indicating that the model acquires reusable manipulation capability rather than merely memorizing task templates.
We also find that the model continues to serve as a stronger prior after fine-tuning: on a 15-task real-robot suite, \walloss{} attains an average task progress of 60.5\%,  surpassing \pimodel{}~\citep{pi05} by 17.5\% and widening the margin to \textbf{26\%} on the 10-task manipulation subset.
Multimodal evaluation further reveals stable overall performance alongside a \textbf{21.8\%} gain on embodied grounding, indicating that action training strengthens embodied perception without eroding general vision-language competence.

The core methodological contribution underlying these results is \textbf{gradient-bridged co-training}, which proves decisive for real-robot performance. The starting point is the tension inherent to VLA training. Continuous flow matching~\citep{lipman2022flow} constitutes the natural execution interface, as it models unquantized robot actions directly; yet on its own, it only weakly updates the pretrained VLM backbone. Discrete action-token prediction exhibits the complementary property: next-token cross-entropy is native to the VLM training interface and strongly shapes the backbone, but decoded discrete actions are too coarse for precise control. Freezing or truncating gradients preserves the VLM prior, but at the cost of preventing precise action targets from shaping the large pretrained backbone.

\walloss{} resolves this tension through gradient-bridged co-training. During pretraining, action-token cross-entropy supplies the gradient bridge: a strong, VLM-native action signal that updates the backbone in a direction aligned with continuous control. In strength, it drives backbone updates far more effectively than flow matching, by virtue of sharing the autoregressive interface of VLM pretraining; in alignment, its gradient direction remains positively correlated with that of flow matching, shaping features that continuous control can subsequently exploit. Multimodal cross-entropy on grounded vision-language data provides the anchor, tying the backbone to instruction following, visual grounding, and embodied scene understanding; its update direction is largely orthogonal to action optimization, and therefore complements the bridge rather than competing with it. Flow matching, in turn, trains the continuous action generator used at inference. Gradient analysis confirms why all three components are necessary: beyond the early training stage, flow matching contributes only a small though persistent share of the backbone update, while the dominant updates originate from the two cross-entropy losses. Ablating the gradient bridge, isolating gradients, or delaying co-training substantially degrades downstream behavior (\Cref{sec:ablation_strategy}).

This recipe is instantiated through three design choices.
First, a \textbf{Mixture-of-Transformers (MoT) backbone}~\citep{liang2024mixture} separates a \textit{VL Expert} and an \textit{Action Expert} at every layer.
Vision, language, and discrete action tokens are routed through the VL Expert, whereas continuous action signals are routed through the Action Expert. Crucially, gradients still flow end-to-end across both experts, so the split amounts to a routing decomposition rather than gradient isolation.
Second, a \textbf{Vision-Aligned RVQ~\citep{lee2022autoregressive} Action Tokenizer} replaces rule-based FAST tokenization~\citep{fast}, rendering the discrete tokens a more semantic training interface for the VLM backbone.
Third, \textbf{Action-Space Supervision}~\citep{feng2026demystifying} strengthens flow matching by imposing the loss directly in the raw action space, rather than on the velocity field as in standard flow matching, thereby accelerating convergence and stabilizing continuous action generation.

\walloss{} is trained in a single stage on a three-source mixture: high-quality self-collected manipulation data, a curated set of open-source multi-embodiment trajectories~\citep{oxe,droid,robomind,robomind2,agibot_world,bridgev2,rt1}, and a 90M-sample multimodal corpus that includes embodied bridge samples synthesized from action trajectories.
This mixture mirrors the training design: robot trajectories teach execution, multimodal samples preserve grounded understanding, and embodied bridge samples connect the two.

Beyond the modeling recipe, we engineer a deployment-grade inference stack that combines CUDA Graph capture and custom fused kernels, achieves a 4× end-to-end speedup over a PyTorch eager-mode baseline, and sustains real-time control (15 Hz) at high input resolution.

We summarize our contributions as follows:
\begin{itemize}[leftmargin=*,itemsep=2pt]
    \item \textbf{A pretrained VLA that can be evaluated directly on robots.}
    We release an open-source 4B VLA, built upon a 3B VLM backbone, whose pretrained checkpoint is evaluated on a 17-task real-robot zero-shot suite prior to any task-specific fine-tuning, with 15 Hz inference speed at high input resolution.

    \item \textbf{A practical co-training recipe for action-aware VLM backbones.}
    We identify action-token cross-entropy as the gradient bridge, multimodal cross-entropy as the grounding anchor, and flow matching as the continuous execution objective, and show that the three are jointly necessary.

    \item \textbf{An integrated architecture and training design.}
    We combine an MoT routing backbone, a Vision-Aligned RVQ Action Tokenizer, and Action-Space Supervision to instantiate this recipe at scale, while leaving the deployment-time action interface unchanged.

    \item \textbf{A multi-axis evaluation of pretrained and adapted capability.}
    We evaluate out-of-the-box real-robot behavior, post-finetuning real-robot performance and multimodal capability, demonstrating that strong pretraining yields both direct real-robot capability and improved downstream adaptation.
\end{itemize}

The remainder of this report is organized as follows: \Cref{sec:methods} describes the model and training recipe, \Cref{sec:data} details the data pipeline, \Cref{sec:experiments} presents the main experiment results, \Cref{sec:ablation} validates the design choices, \Cref{sec:related} discusses related work, and \Cref{sec:discussion} concludes with limitations and future directions.

%% file: sec/2_methods.tex
\section{Methods}\label{sec:methods}

\begin{figure}[t]
    \centering
    \includegraphics[width=1.0\linewidth]{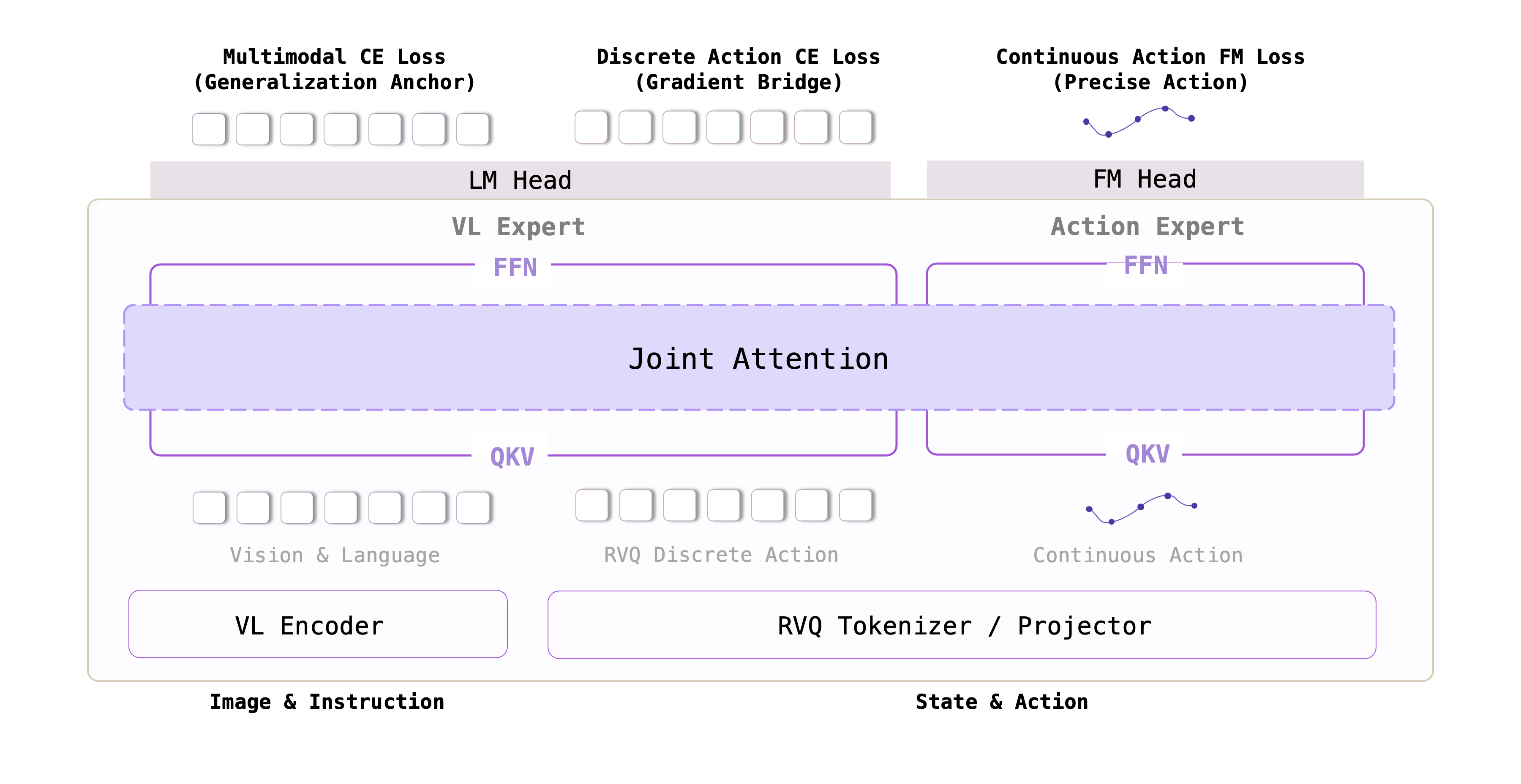}
    \caption{\textbf{Gradient-bridged co-training and MoT routing in \walloss{}.}
    Three complementary objectives shape the pretrained policy: multimodal cross-entropy (CE) preserves grounded vision-language knowledge, action-token CE provides the gradient bridge that adapts the VLM backbone toward control, and flow matching (FM) supervises the continuous actions used for deployment.
    The Mixture-of-Transformers architecture routes vision-language tokens through the VL Expert and continuous-action computation through the Action Expert, with joint attention enabling end-to-end gradient flow between the two experts.}
    \label{fig:method_main}
\end{figure}
In \walloss{},
pretraining is accordingly organized around three types of data modalities:
discrete action tokens shape the VLM backbone during training, continuous flow matching supplies the action output at inference and grounded multimodal data preserves the vision-language prior.
This section describes the architecture, tokenizer, training losses, and action interface used to realize this design in a single training stage.

\subsection{Architecture}\label{sec:architecture}

\subsubsection{Backbone Routing}\label{sec:overall_arch}

\walloss{} is initialized from Qwen2.5-VL-3B-Instruct~\citep{Qwen2.5-VL} and extended with a Mixture-of-Transformers (MoT) backbone, yielding a model with over 4B parameters.
The original 3B VLM is retained as the \emph{VL Expert}, while the added \emph{Action Expert}, together with action projections in continuous-action heads, provides the additional action-generation capacity.
Four token streams, vision, language, proprioception, and discrete actions, are routed through the VL Expert,
while noisy continuous action tokens are routed through the Action Expert, which is trained for flow-matching action generation.

This split constitutes a routing decomposition rather than a gradient-stopping mechanism.
The two experts share the same sequence-level attention context, allowing the Action Expert to attend to visual and linguistic information when generating continuous actions.
The attention mask renders discrete and continuous action tokens mutually invisible in the forward pass, which permits the two action pathways to be trained and evaluated independently. In the meantime, gradients are not blocked from flow matching to VL Expert through shared attention.

\subsubsection{Vision-Aligned RVQ Action Tokenizer}\label{sec:vq_tokenizer}

We adopt discrete action tokens because next-token cross-entropy is the training interface most directly compatible with the VLM backbone.
The tokenizer must therefore expose structured action semantics for backbone training~\cite{clip,li2024decisionnce,he2020momentum,zhang2026clapcontrastivelatentaction}, not merely achieve low-distortion reconstruction.
Accordingly, we replace the FAST tokenizer~\citep{fast} with a learned Vision-Aligned Residual Vector Quantization~(RVQ) Action Tokenizer, trained on heterogeneous robot data spanning diverse embodiments.

\begin{figure}[t]
    \centering
    \includegraphics[width=1\linewidth]{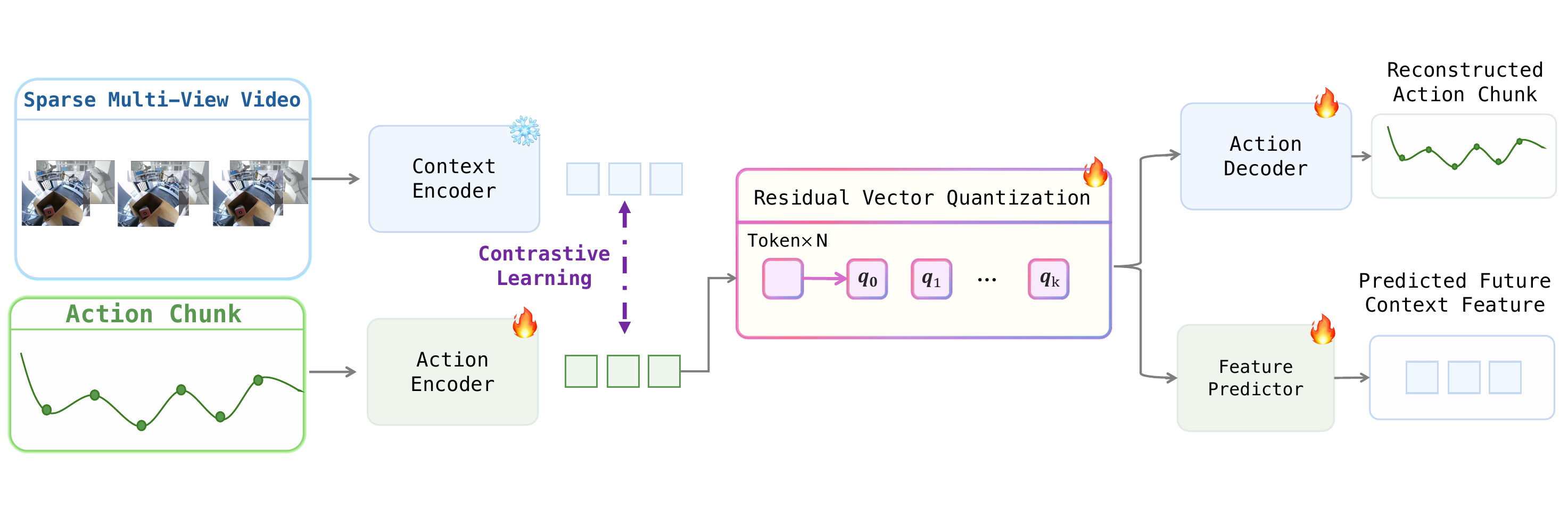}
    \caption{\textbf{Overview of Vision-Aligned RVQ Action Tokenizer.} Our framework compresses observation-conditioned delta action sequences into multi-level discrete tokens via residual vector quantization. By incorporating auxiliary visual-action and future-observation objectives, the tokenizer serves as a semantic training interface for the VLM backbone rather than a mere action compressor.}
    \label{fig:tokenizer_report}
\end{figure}

The tokenizer operates in the delta-action space and follows an Encoder–RVQ–Decoder structure, as shown in \Cref{fig:tokenizer_report}.
The encoder compresses observation-conditioned action chunks via temporal cross-attention; the RVQ codebooks capture coarse motion structure at early levels and fine residual corrections at later levels; and the decoder reconstructs action sequences conditioned on observation states.
Beyond reconstruction, three objectives jointly shape the token space: visual-action alignment pulls action latents toward VLM visual features, next-frame prediction encourages tokens to encode action consequences, and DCT-domain reconstruction suppresses high-frequency jitter.
The resulting discrete action representation is simultaneously reconstructable, visually aligned, and physically smooth.

\subsection{Training Recipe}\label{sec:training}

\subsubsection{Gradient-Bridged Co-Training}\label{sec:cotrain}

During pretraining, we jointly optimize three objectives in a single stage.
First, action-token cross-entropy predicts the RVQ action tokens autoregressively and serves as the gradient bridge between \emph{VL Encoder} and \emph{Action Expert}: a strong, VLM-native signal that renders the backbone action-aware.
Second, multimodal cross-entropy loss trained on \emph{VL Encoder} on grounded vision-language samples serves as the anchor that preserves instruction following, visual grounding, and embodied scene understanding.
Third, flow matching trains the \emph{Action Expert} to generate the continuous action chunks used at inference.
We illustrate the model architecture of the gradient bridge in \Cref{fig:method_main}. The discrete pathway is therefore primarily a training-time pathway, whereas the continuous pathway is the deployment-time pathway.

This design is motivated by the gradient dynamics analyzed in our ablations.
Flow matching defines the execution objective, but beyond the early training stage its contribution to the VLM-backbone update stabilizes at a small but persistent share of roughly 5\%,
the dominant backbone updates instead originate from the two cross-entropy losses.
Action-token cross-entropy is aligned with action generation and biases the backbone toward control-relevant features, while multimodal cross-entropy anchors the update in the original vision-language prior.
Accordingly, \walloss{} preserves end-to-end gradient flow rather than adopting the stop-gradient design of \pimodel{}~\citep{pi05}: the small flow-matching residual still matters for action quality, while action-token cross-entropy carries most of the backbone shaping.

In summary, \walloss{} is trained with the following composite objective:
\begin{equation}\label{eq:total_loss}
    \mathcal{L} =
    \mathcal{L}_{\text{flow}}
    + \lambda_{\text{act}} \cdot \mathcal{L}_{\text{act-CE}}
    + \lambda_{\text{mm}} \cdot \mathcal{L}_{\text{mm-CE}},
\end{equation}
where $\mathcal{L}_{\text{act-CE}}$ denotes the autoregressive RVQ action-token loss, $\mathcal{L}_{\text{mm-CE}}$ denotes the multimodal next-token loss, and $\lambda_{\text{act}}=\lambda_{\text{mm}}=0.01$.
Empirically, $\mathcal{L}_{\text{flow}}$ is roughly two orders of magnitude smaller than the cross-entropy terms under Action-Space Supervision; the shared weight of $0.01$ places the cross-entropy losses on a scale comparable to the flow-matching loss, preventing language-style prediction from dominating action learning.
The relative contribution of action-token and multimodal cross-entropy is in turn controlled by batch composition, with action and multimodal data mixed at a $9{:}1$ ratio.
At inference, decoding defaults to the continuous flow-matching pathway, which is decoupled from the discrete pathway by attention masking (\Cref{sec:overall_arch}); the discrete pathway's role is confined to carrying the gradient bridge during training.

\subsubsection{Action-Space Supervision}\label{sec:action_space}

Flow matching starts from a noisy action chunk and learns a velocity field that transports noise to the clean action. We use the linear Gaussian probability path
\begin{equation}\label{eq:noisy_action}
    \mathbf{A}_t^\tau = \tau \mathbf{A}_t + (1-\tau)\epsilon, \quad \epsilon \sim \mathcal{N}(\mathbf{0}, \mathbf{I}),
\end{equation}
where $\tau=0$ corresponds to pure noise and $\tau=1$ to the clean action.
Following \pizero{}~\citep{pi0}, we bias timestep sampling toward the high-noise regime:
\begin{equation}\label{eq:beta_sampling}
    u \sim \mathrm{Beta}(1.5, 1), \qquad \tau = s(1-u), \quad s = 0.999.
\end{equation}
Since $\tau=0$ corresponds to pure noise, this transformation concentrates probability mass on high-noise timesteps.

Robot action sequences are low-dimensional and smooth: their task-relevant structure resides primarily in the low-frequency trajectory shape rather than in high-frequency detail.
We accordingly hypothesize that the high-noise regime is particularly critical for recovering the global shape of robot trajectories, whereas low-noise steps serve mainly to refine residual detail.
Unlike natural images, in which both high- and low-frequency components carry rich semantics, robot actions tend to concentrate useful task structure in smooth trajectory trends; supervision quality in the high-noise regime therefore largely determines the generation ceiling.
Guided by this intuition, we retain velocity prediction as the network output but define the loss on the recovered action:
\begin{equation}\label{eq:pred_action}
    \hat{A} = A^\tau + (1-\tau)\cdot f_\theta(A^\tau, \tau),
\end{equation}
\begin{equation}\label{eq:action_loss}
    \mathcal{L}_A = \mathbb{E}_{\tau, \epsilon}\left[\left\|\hat{A} - A\right\|^2\right].
\end{equation}
This formulation is equivalent to a $(1-\tau)^2$-weighted loss in velocity space:
\begin{equation}\label{eq:equiv_loss}
    \mathcal{L}_A = \mathbb{E}_{\tau, \epsilon}\left[(1-\tau)^2\left\|f_\theta(A^\tau, \tau) - (A-\epsilon)\right\|^2\right].
\end{equation}
The induced weighting emphasizes high-noise steps, at which the global action trajectory is formed, consistent with the trajectory-smoothness hypothesis above.
This data-space formulation is related to x-prediction in diffusion models~\citep{li2025back}; here, however, the motivation derives from the spectral structure of robot actions rather than from variance considerations.
The controlled ablation in \Cref{sec:ablation_flowloss} supports this design choice, demonstrating that Action-Space Supervision improves convergence speed, peak performance, and training stability.

\subsubsection{Action Interface}\label{sec:io_format}

The model follows a VLM-style conversation sequence.
For action prediction, the input takes the form:
\begin{center}
\small
\texttt{[System]} Embodiment prompt \texttt{[User]} Observation: \{camera\}: $\langle$image tokens$\rangle$ \ldots\ Instruction: \{task\} Proprioception: $\langle$proprio tokens$\rangle$ \texttt{[Assistant]} $\langle$action\_ar\_token$\rangle$ $\langle$action\_flow\_token$\rangle$ $\times N$
\end{center}
Here $\langle$action\_ar\_token$\rangle$ denotes the discrete RVQ action tokens consumed by the autoregressive cross-entropy pathway, while $\langle$action\_flow\_token$\rangle$ denotes the continuous-action query tokens consumed by the flow-matching pathway; $N$ is the number of frames in the predicted action horizon.
For multimodal understanding samples, the output is text-only and contains no action tokens.
The model supports an arbitrary number of camera views, sampled according to the data source.
Each task is annotated with both a goal-level instruction (e.g., "tidy up the desk") and step-level sub-instructions (e.g., "throw it into the trash can"); during training, one granularity is sampled per step, allowing the model to follow instructions at multiple levels of abstraction.
Paraphrases are additionally used to reduce overfitting to surface wording.
Proprioception is discretized into text-numerical tokens and randomly dropped or perturbed during training, reducing dependence on noise-free robot state.
At inference, discrete tokens are not decoded into executable actions; the Action Expert generates continuous action chunks through multi-step flow-matching denoising, with the two pathways kept decoupled in the forward pass by attention masking.

We adopt relative action representations~\citep{feng2026demystifying} and 6D rotations~\citep{hempel20226d} rather than Euler angles or quaternions, in order to avoid the discontinuities and ambiguities of SO(3).
The resulting model action space is 26-dimensional: for each arm, a relative 3D position, a relative 6D rotation (both relative to the current end-effector pose), and a 1D gripper state (20D in total), plus a 3D mobile base velocity, a 1D lift height, and 2D head actuation.
Both discrete and continuous pathways predict a one-second action horizon, with the number of frames adjusted to the control frequency of each data source.

\subsubsection{Optimization and Configuration}\label{sec:muon}\label{sec:config}
We empirically find that Muon~\citep{liu2025muon} is effective in our co-training setup, yielding both faster convergence and improved training stability.
Muon is applied to the 2D parameters of each expert, while AdamW~\citep{loshchilov2017decoupled} handles the visual embeddings and LM head.
Muon orthogonalizes the momentum prior to applying it as the update, producing spectrally normalized updates that are invariant to gradient magnitude. This invariance is critical in our setting, where the Action Expert and the VL Expert exhibit markedly different gradient scales.

To make Muon practical at scale, we implement DMuon, a distributed Muon runtime compatible with hybrid-parallel training. DMuon assigns each matrix parameter to a unique owner rank using Longest-Processing-Time-first load balancing, performs the matrix-level Newton–Schulz update once on the owner, and synchronizes the updated matrix state through reduce-to-owner communication followed by asynchronous post-step broadcast. The broadcast is issued on a dedicated CUDA stream and overlapped with the next forward pass whenever possible. DMuon also accelerates owner-local matrix-level Newton–Schulz computation with symmetry-aware kernels and shape-aware autotuning. In our training stack, DMuon reduces the additional optimizer overhead introduced by Muon from roughly (2$\times$) the forward-plus-backward time in a naive implementation to about (0.02$\times$), an approximately (100$\times$) reduction.

Our pretraining uses an effective global batch size of 8192, bf16 mixed-precision training, gradient clipping at 1.0, a cosine learning-rate schedule with linear warmup, and a peak learning rate of $1\times10^{-4}$.

Images are resized with aspect ratio preserved and the longer side fixed at 448 pixels;
stationary frames in all pretrained datasets are filtered out to prevent overfitting to idle actions.
Fine-tuning uses a learning rate of $5\times10^{-5}$ with all modules trainable, and retains the same joint discrete-plus-continuous objective; as shown in \Cref{sec:ablation_strategy}, this co-training setup yields a large relative gain over flow-only fine-tuning.

\subsection{Inference Optimization}

Real-time control is a defining requirement of VLA models deployed on physical robots. Unlike offline vision-language tasks, manipulation policies operate in closed loop with the environment, where every additional millisecond of inference latency translates into delayed actuation, degraded tracking of dynamic targets, and ultimately lower task success rates. A control frequency on the order of tens of Hertz is therefore a functional prerequisite rather than a performance luxury. This requirement is at odds with the architectural trend of modern VLAs, which couple a large vision encoder, a billion-scale language backbone, and an iterative action decoder into a single forward pass. The cost is further amplified when high-resolution visual input is required for fine-grained manipulation: ViT~\cite{dosovitskiy2020image} compute grows quadratically with the number of image tokens, and the resulting longer visual context inflates the key-value (KV) cache of the language backbone, lengthening every attention operation downstream. Inference optimization for high-resolution VLA models is therefore substantially harder than for their low-resolution counterparts, and warrants dedicated treatment rather than off-the-shelf acceleration.

We optimize the inference pipeline of \walloss{} along two complementary axes, each targeting a dominant source of overhead identified through profiling. The denoising step is a memory-bound workload: GPU kernel execution time is shorter than the CPU launch latency, making CPU dispatch the bottleneck. The GPU consequently stalls between kernels, and the resulting bubbles, rather than compute, dominate end-to-end latency. Since the per-step computation graph is static, we capture the entire denoising step as a single CUDA Graph, which removes CPU dispatch from the critical path and eliminates these bubbles. Beyond GEMM and attention~\citep{vaswani2017attention}, the remaining operators (RoPE~\cite{su2024roformer}, RMSNorm~\citep{zhang2019root}, and similar elementwise and reduction patterns) are individually inexpensive but collectively incur substantial HBM traffic when executed as separate PyTorch ops, each of which materializes intermediate tensors. We fuse these operators into monolithic custom CUDA kernels that perform end-to-end computation in registers, thereby eliminating intermediate materialization and yielding 2–10× speedups over the native implementation.

\paragraph{Result.} We evaluate the optimized inference stack on a single RTX 5090 with three-view image input at 224$\times$224 and 448$\times$448 resolutions. Our optimized implementation reaches approximately 21 Hz and 15 Hz at the two resolutions respectively with the standard denoising step $T=10$ setting, corresponding to a 4$\times$ end-to-end speedup compared to a PyTorch eager-mode baseline.  The relative speedup remains substantial at 448$\times$448, where the baseline is most constrained by the quadratic cost of ViT and the inflated KV cache, confirming that the proposed optimizations are particularly effective in the high-resolution regime required by real-world manipulation. These numbers represent only a partial exploration of the optimization space; further gains are left to future work.

%% file: sec/3_data.tex
\section{Data Recipe and Management}\label{sec:data}

The pretraining dataset for \walloss{} combines diverse, high-quality self-collected manipulation data with open-source multi-embodiment datasets and targeted multimodal corpora. The resulting mixture is balanced across task diversity, embodiment coverage, and data-source distribution, providing a strong foundation for learning robust manipulation behavior during pretraining.

\subsection{Robot Manipulation Data}\label{sec:action_data}

We unify action representations across embodiments for both self-collected and open-source data. Specifically, we map each platform's actions to a shared semantic schema (bimanual end-effector poses, joint positions, gripper states, mobile-base motion, lift/waist actuation, and head actuation). We also synchronize camera streams by timestamp within each platform, while keeping platform-specific camera counts and viewpoints unchanged. Details of this cross-source preprocessing pipeline are provided in \Cref{sec:data_preprocessing}.

\subsubsection{Self-Collected Robot Manipulation Data}\label{sec:inhouse_data}

Our self-collected data forms the core of the manipulation pretraining corpus and covers thousands of distinct tasks.

\paragraph{Embodiment composition.} Our self-collected data comes from two main platform types: tabletop bimanual systems and mobile manipulators, covering both fixed-workspace tasks and mobile manipulation scenarios. We further enrich these robot-platform data sources with an embodiment-free collection device.

In addition to teleoperated whole-robot platforms, we introduce XRZero-G0 \cite{wang2026xrzero}, a proprietary embodiment-free device that enables low-cost data collection across diverse scenes without tying the data to a specific robot morphology, thereby expanding both environmental and task diversity.

\paragraph{Scene and task distribution.} The self-collected corpus covers diverse scenes and tasks. Scenes include unstructured real-world environments (household, industrial, and office), which capture natural clutter and lighting variation, and controlled collection rooms, which improve reproducibility and cross-batch consistency. Tasks are labeled along three complementary axes---\emph{manipulation complexity}, \emph{trajectory duration}, and \emph{special attributes} (e.g., spatial reasoning and deformable-object interaction)---covering both industry-relevant operations and everyday behaviors. We use these task labels to balance the training mixture (\Cref{sec:data_preprocessing}), so that a small number of high-volume tasks do not dominate the corpus.

\paragraph{Task segmentation and language annotation.} At the episode level, we annotate each trajectory with two complementary instruction types: a brief instruction that specifies only the final goal, and a detailed instruction that explicitly lists intermediate sub-goals and key action transitions in temporal order. We then expand both instruction types using large language models to improve linguistic coverage while preserving semantics. In parallel, we split long teleoperated trajectories into temporally coherent sub-trajectories, each corresponding to an atomic sub-goal, and annotate each segment with a matched instruction. Together, episode-level brief/detailed annotation and segment-level supervision support both global goal understanding and fine-grained cross-modal alignment, yielding thousands of base instructions and tens of thousands of paraphrases.

\subsubsection{Open-Source Multi-Embodiment Data}\label{sec:opensource_data}

\begin{figure}[t]
  \centering
  \includegraphics[width=\textwidth]{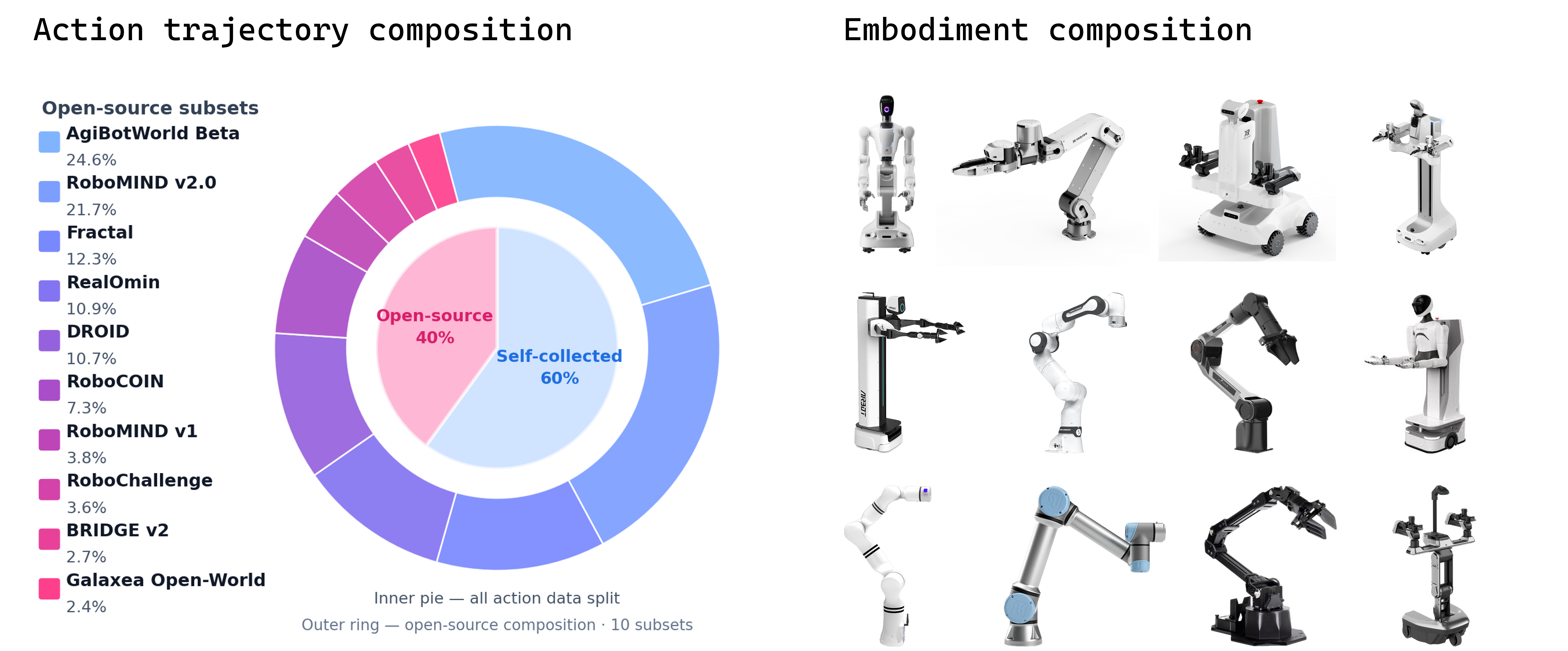}
  \caption{Robot manipulation data landscape used for pretraining. Left: retained dataset subsets and trajectory composition, including self-collected robot manipulation data, RoboMIND v1~\citep{robomind}, RoboMIND v2.0~\citep{robomind2}, AgiBotWorld Beta~\citep{agibot_world}, RoboCOIN~\citep{robocoin}, RoboChallenge~\citep{robochallenge}, Galaxea Open-World~\citep{galaxea_openworld}, RealOmin~\citep{realomin}, DROID~\citep{droid}, BRIDGE v2~\citep{bridgev2}, and Fractal/Google Robot~\citep{rt1}. Right: embodiment composition summarizing the morphology-level diversity contributed by self-collected data and open-source subsets.}
  \label{fig:opensource_data_landscape}
\end{figure}

Open-source manipulation data extends embodiment and scene coverage without requiring additional self-collected data collection. We perform unified curation on a selection of high-quality open-source datasets (format alignment, metadata verification, cross-source field mapping) and filter samples by timestamp continuity, action-observation plausibility, and frame anomalies. \Cref{fig:opensource_data_landscape} visualizes the 10 retained subsets and the resulting embodiment coverage, highlighting the broad scale and morphological diversity of the curated open-source corpus.

\subsubsection{Data Preprocessing and Sampling Strategy}\label{sec:data_preprocessing}

Multi-source robot data exhibits significant heterogeneity in quality, format, and action-space representation: different datasets use varying storage formats and field naming conventions, and even the same embodiment may have inconsistent action definitions across datasets (e.g., coordinate-frame orientation, rotation representation conventions, gripper-state polarity). If left unaddressed, these differences directly introduce noise and degrade the effectiveness of joint training across sources. We therefore apply a systematic preprocessing pipeline to all datasets with action annotations.

\paragraph{Action space unification.} We standardize each source into a unified action schema covering bimanual end-effector poses, joint positions, gripper states, mobile base motion, lift/waist actuation, and head motion. For datasets that provide only joint states, we recover end-effector poses through forward kinematics using the corresponding platform URDF. We further normalize the physical semantics of actions across embodiments: $x$ points forward, $y$ left, and $z$ upward; the zero-rotation pose corresponds to a forward-facing gripper with a horizontal opening; and larger gripper values denote wider opening. When source annotations use Euler angles, we retain them only at the normalized source level and convert rotations to the 6D model representation before training to avoid discontinuities and gimbal lock.

\paragraph{Video--action temporal alignment.} Different sensors often have inconsistent sampling frequencies and timestamp precision. We perform unified temporal alignment of video frames and action/state frames across all robot manipulation data by timestamp. For cases where video and action frames cannot be strictly matched one-to-one, we use a nearest-timestamp strategy: for each video frame, the state frame with the closest timestamp is retrieved as the corresponding observation--action pair.

\paragraph{Erroneous data repair.} Quality inspection further flags typical recording errors---such as swapped camera mappings or anomalous end-effector poses---which we correct where possible (e.g., remapping cameras or recovering poses via forward kinematics) and discard otherwise.

\paragraph{Stationary frame filtering.} Near-stationary frames---identical observations paired with both near-zero and non-zero actions---introduce supervision noise and induce idle pauses at inference. We filter them by normalizing actions with global statistics and removing frames whose per-dimension difference from the last retained frame falls below a threshold. Beyond cleaner gradient signal, we observe visibly more compact task execution cadence post-training (reduced redundant pauses).

\paragraph{Data sampling strategy.} Multi-source data is imbalanced at two levels: inter-source (trajectory counts differ by orders of magnitude between self-collected and open-source subsets) and intra-source (per-task long-tail within each dataset). Proportional sampling would let high-frequency tasks dominate each epoch's gradient and leave rare tasks underrepresented. We therefore define sampling groups using source and task labels, then apply power sampling within the resulting groups: for the $i$-th group with $n_i$ trajectories, sampling weight $w_i = n_i^p$ and normalized count $s_i = N \cdot w_i / \sum_j w_j$, where $N$ is the total trajectory budget per epoch. We set $p = 0.5$ (square-root sampling), boosting small-group frequency while preserving the statistical weight of large groups. A per-group cap with iterative reallocation prevents very large groups from monopolizing the budget. After sampling, one training epoch comprises over one million trajectories (${\sim}60\%$ self-collected, ${\sim}40\%$ open-source).

\subsection{Multimodal Data}\label{sec:multimodal_data}

Multimodal data provides the generality anchor in gradient-bridged co-training: it carries the multimodal CE loss that keeps the VLM backbone tied to grounded vision-language understanding while action-token CE shapes it for control. Beyond generic vision-language alignment, the corpus supervises object recognition, scene composition, spatial relations, affordance, and task-relevant interaction cues, all of which help the model transfer to unseen scenes, object instances, and instructions. The multimodal corpus totals approximately 90 million samples, consisting of 78 million open-source samples and 12 million embodied bridge samples constructed directly from action trajectories. During training, action and multimodal samples are mixed at a 9:1 batch-sampling ratio: multimodal samples are trained with autoregressive next-token CE, while action trajectories jointly provide action-token CE and flow-matching supervision.

\begin{table}[H]
  \centering
  \caption{Open-source multimodal data organized by category.}
  \label{tab:multimodal_opensource}
  \small
  \setlength{\tabcolsep}{5pt}
  \renewcommand{\arraystretch}{1.18}
  \begin{tabularx}{0.98\textwidth}{@{}>{\raggedright\arraybackslash}p{0.20\textwidth}
      >{\raggedright\arraybackslash}p{0.24\textwidth}
      >{\raggedright\arraybackslash}X@{}}
    \toprule
    \textbf{Category} & \textbf{Task Types} & \textbf{Datasets} \\
    \midrule
    General vision-language & Captioning, VQA, pointing, reasoning & CAPSFUSION~\citep{CapsFusion}, Cambrian~\citep{Cambrian}, PixMo-Cap~\citep{molmo2024}, COCO~\citep{COCO}, VQAv2~\citep{VQAv2}, PixMo-Point~\citep{molmo2024}, OneThinker~\citep{feng2025onethinker} \\
    \addlinespace
    Embodied perception & Grounding, pointing, spatial, affordance & RoboPoint~\citep{RoboPoint}, SpaceThinker~\citep{spacethinker2025}, OpenSpaces~\citep{openspaces2025}, SpaceOm~\citep{spaceom2025}, RefSpatial~\citep{refspatial2025}, CrossPoint~\citep{crosspoint2025}, SenseNova-SI~\citep{sensenovasi2025}
    \\
    \addlinespace
    Embodied cognition & Task VQA, interaction, long-horizon reasoning & Robo2VLM~\citep{Robo2VLM}, EO-Data~\citep{eo1_2025}, RoboVQA~\citep{robovqa2023}, Cosmos-Reason1~\citep{cosmosreason2025} \\
    \bottomrule
  \end{tabularx}
\end{table}

\begin{figure}[H]
  \centering
  \includegraphics[width=\linewidth,height=0.52\textheight,keepaspectratio]{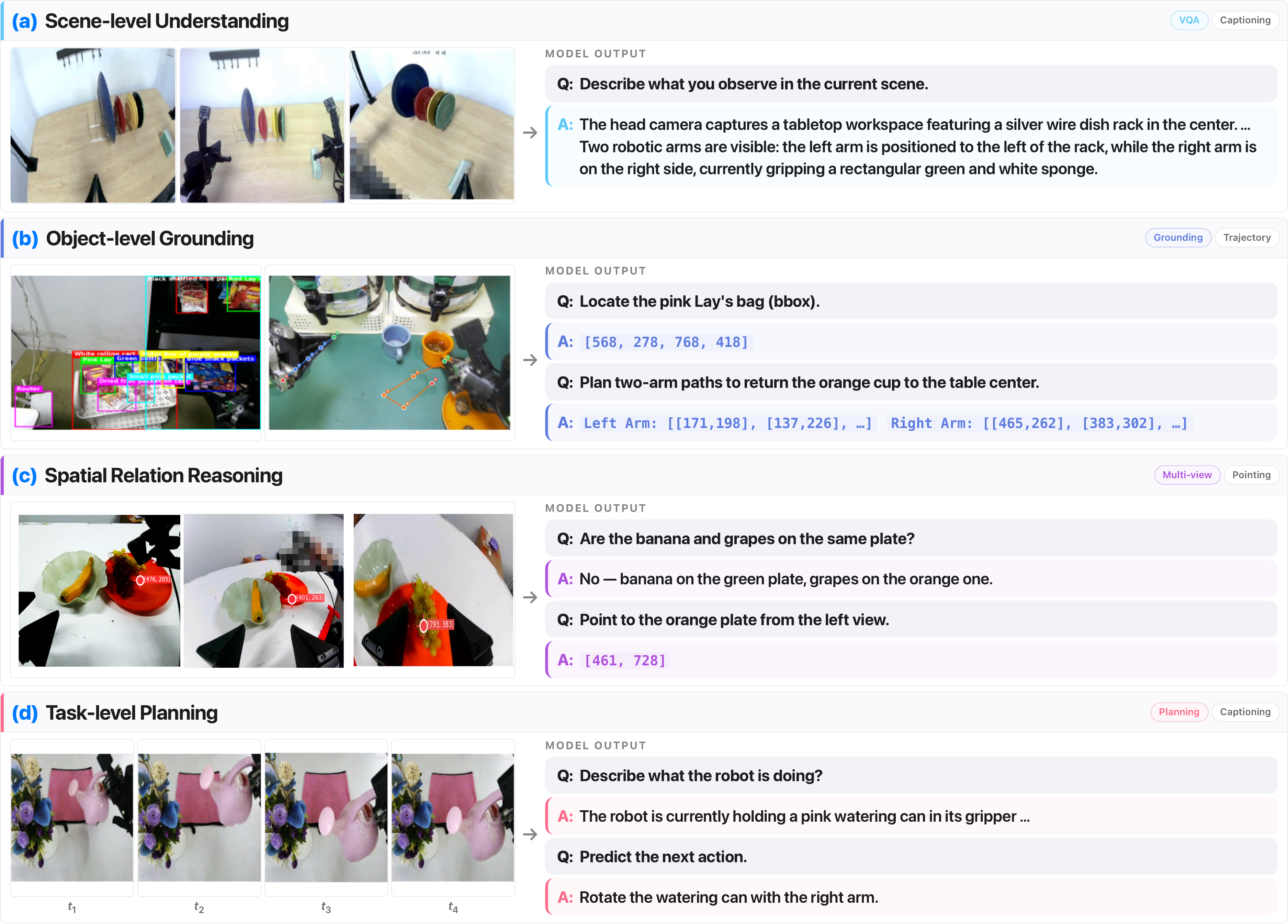}
  \caption{Embodied bridge data construction and task taxonomy. Bridge samples are generated from action trajectories and organized into object, spatial, scene, and task understanding objectives, aligning multimodal supervision with executable robot behavior.}
  \label{fig:bridge_data}
\end{figure}

\paragraph{Open-source multimodal data.} The open-source portion serves two complementary goals---maintaining general VLM capabilities and injecting embodiment-relevant understanding---and is instantiated as three dataset categories. \emph{General vision-language data} provides broad supervision for captioning, general question answering, pointing, and open-ended reasoning; \emph{embodied perception data} targets object grounding, affordance understanding, and spatial reasoning; and \emph{embodied cognition data} covers task VQA, interaction understanding, and long-horizon reasoning. \Cref{tab:multimodal_opensource} lists representative datasets for each category. We apply strict quality control to open-source data, combining model-based verification with human annotation to filter inaccurate, low-information, and excessively repetitive samples.

\paragraph{Embodied bridge data.} The second component, which we term \emph{embodied bridge data}, is constructed through an automated data pipeline directly from the action pretraining corpus. We use the term "bridge" because these samples provide an explicit connection from multimodal understanding to action prediction---they originate from the same trajectories, observations, and task contexts as action learning, and are thus natively aligned with executable robot behavior.

As shown in \Cref{fig:bridge_data}, bridge data is organized along four levels of understanding. \emph{Object understanding} supervises object grounding, pointing, 2D trajectory prediction, and attribute VQA; \emph{spatial understanding} covers inter-object relations and multi-view pointing; \emph{scene understanding} addresses scene captioning and scene-level VQA; and \emph{task understanding} targets task-progress prediction, affordance estimation, and next-step action planning.

For pointing and grounding tasks, dedicated spatial tokens are introduced, unifying all coordinate annotations into text format: \texttt{<box>[x1,y1,x2,y2]</box>} and \texttt{<point>[x,y]</point>}.

%% file: sec/4_experiments.tex
\section{Experiments and Results}\label{sec:experiments}

\subsection{Pretrained Model Zero-Shot Evaluation}\label{sec:zeroshot}

We evaluate the pretrained model without fine-tuning using a real-robot zero-shot suite that spans diverse manipulation dimensions. We evaluate the model on 17 tasks in total: 12 seen tasks drawn from within the pretraining data distribution, and 5 unseen tasks consisting of held-out task configurations that were not collected as identical tasks on the current embodiment, thereby testing generalization and transfer capabilities. Task types span five dimensions: semantic understanding, rigid-object manipulation, deformable-object manipulation, fine-grained manipulation, and long-horizon multi-step manipulation. Each task is scored according to a predefined scoring rubric (task progress, maximum 100, evaluated over 10 trajectories). This metric captures fine-grained progress at the "partial completion" level, making it more suitable than binary success rate for evaluating VLA models' foundational manipulation capabilities.
Multiple milestone checkpoints are sampled during training to track capability evolution trends.

\subsubsection{Results}\label{sec:zeroshot_results}

\Cref{tab:zeroshot} summarizes the average task progress for seen and unseen tasks across milestone checkpoints (per-task detailed results in \Cref{sec:appendix_zeroshot}).

\begin{figure}[t]
    \centering
    \begin{subfigure}[t]{0.48\linewidth}
        \makebox[\linewidth][c]{\includegraphics[width=1.06\linewidth,height=0.68\linewidth]{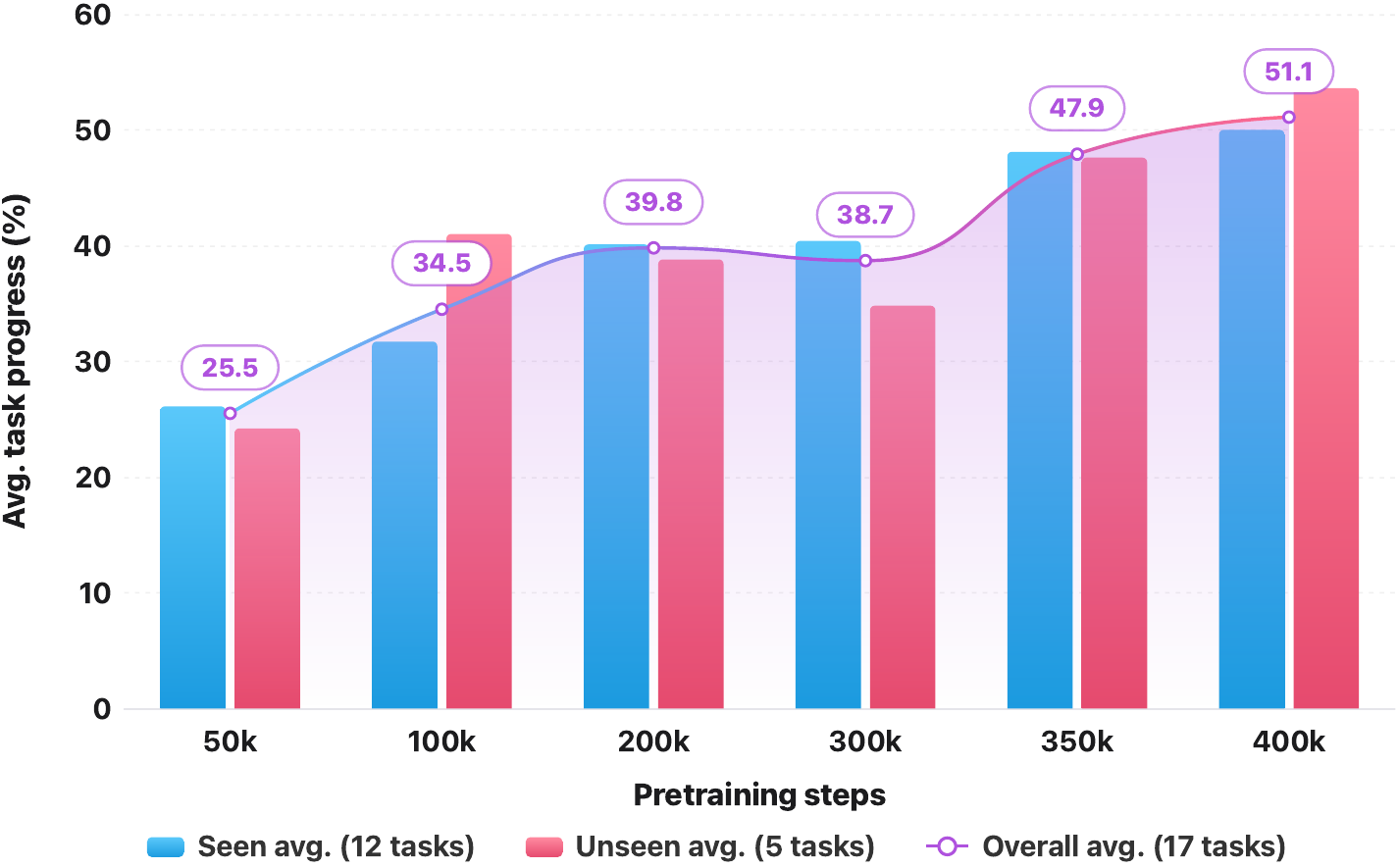}}
        \caption{Seen vs.\ Unseen training trends.}
        \label{fig:zeroshot_trend}
    \end{subfigure}
    \hfill
    \begin{subfigure}[t]{0.48\linewidth}
        \raisebox{0.005\linewidth}{\makebox[\linewidth][c]{\includegraphics[width=1.05\linewidth,height=0.67\linewidth]{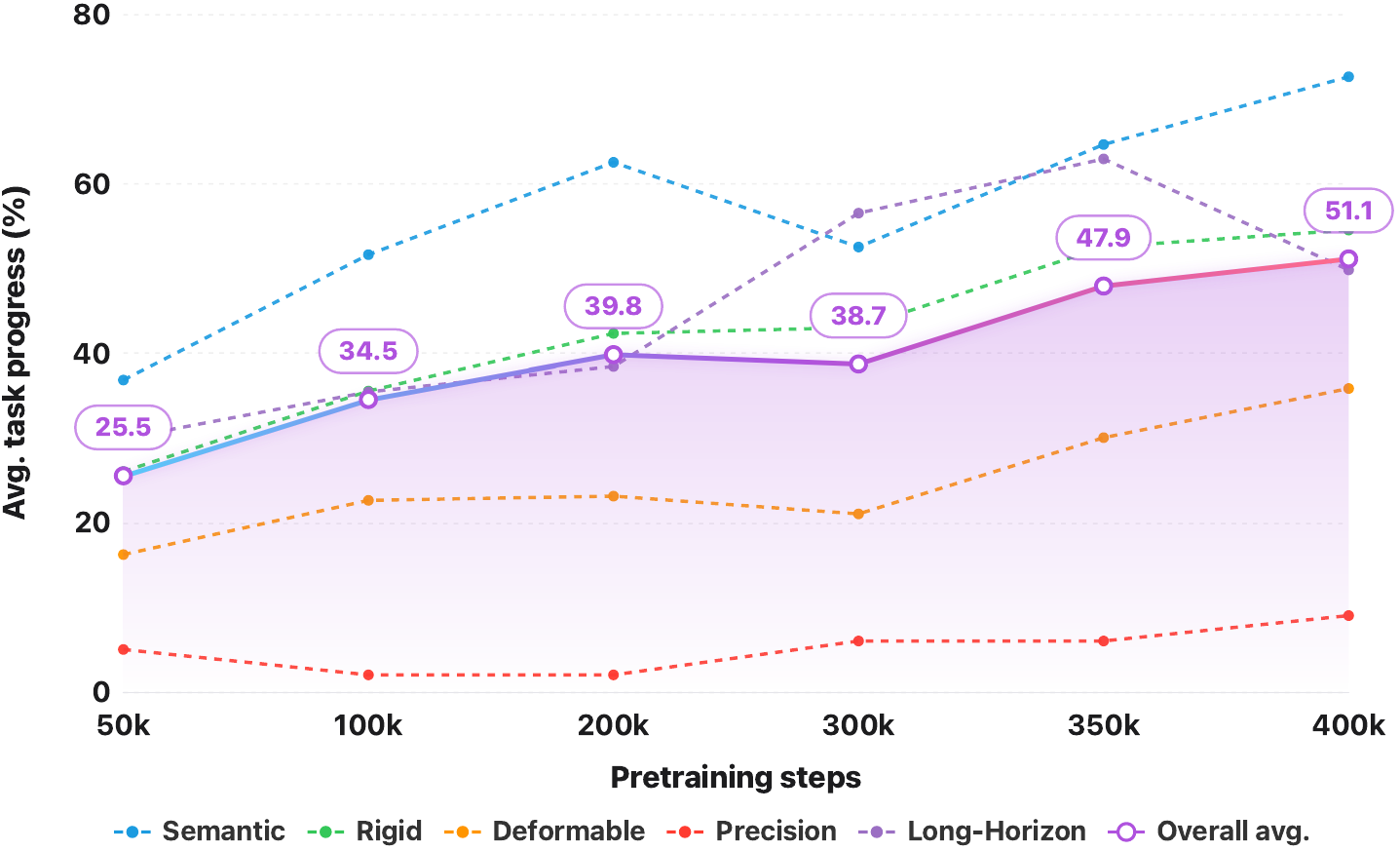}}}
        \caption{Task progress evolution by task category.}
        \label{fig:zeroshot_category}
    \end{subfigure}
    \caption{Zero-shot real-robot evaluation trends across pretraining checkpoints. (a) Average task progress for seen and unseen tasks shows an overall upward trend despite checkpoint-level fluctuations, with held-out tasks reaching 53.6 at the 400k checkpoint. (b) Category-level curves show how capability emerges unevenly across semantic understanding, rigid-object manipulation, deformable-object manipulation, fine-grained manipulation, and long-horizon manipulation, with semantic tasks becoming the strongest zero-shot dimension while precision-demanding categories remain more difficult.}
    \label{fig:zeroshot}
\end{figure}

Before unpacking the aggregate trends, we highlight the strongest pretrained capabilities: at the 400k checkpoint, six tasks already reach or exceed a task-progress threshold of 60\% without any task-specific fine-tuning (\Cref{tab:zeroshot_highlight}), including four tasks above 80\% and two held-out (unseen) tasks. The unseen deformable task \emph{Rope Tightening} reaching 82\% is particularly notable, making pure task-template memorization less likely and suggesting transferable manipulation capability. \Cref{fig:zeroshot_trend} shows the average task progress trends for seen and unseen tasks as training progresses; \Cref{fig:zeroshot_category} further groups tasks by manipulation type, showing the evolution trajectories across five capability dimensions.

\begin{table}[t]
\centering
\caption{Average task progress across pretraining checkpoints for seen and unseen zero-shot evaluation tasks.}
\label{tab:zeroshot}
\setlength{\tabcolsep}{5pt}
\renewcommand{\arraystretch}{1.12}
\begin{tabularx}{0.98\textwidth}{@{}>{\raggedright\arraybackslash}p{0.27\textwidth}
    *{6}{>{\centering\arraybackslash}X}@{}}
\toprule
 & 50k & 100k & 200k & 300k & 350k & 400k \\
\midrule
Seen avg.\ (12 tasks) & 26.1 & 31.7 & 40.1 & 40.4 & 48.1 & \textbf{50.0} \\
Unseen avg.\ (5 tasks) & 24.2 & 41.0 & 38.8 & 34.8 & 47.6 & \textbf{53.6} \\
Overall avg.\ (17 tasks) & 25.5 & 34.5 & 39.8 & 38.7 & 47.9 & \textbf{51.1} \\
\bottomrule
\end{tabularx}
\end{table}

\subsubsection{Analysis}\label{sec:zeroshot_analysis}

From the above experimental results, we summarize the following core findings:

\paragraph{Semantic grounding transfers to action.} From \Cref{fig:zeroshot_category}, semantic understanding tasks are the model's strongest dimension (400k average task progress of 72.6\%), exceeding other categories. Several high-performing seen tasks, including Block Sorting (100\%), Ring Stacking (86\%), and Fruit Sorting (96\%), require the model to ground visual-semantic concepts such as "which color corresponds to which position" or "the ring should be placed on the pole" before executing the manipulation. Among unseen tasks, Toy Basket Placement reaches 58\% at 400k and peaks at 72\% during pretraining, suggesting that VLM-derived semantic knowledge is being transferred to action execution while some held-out tasks remain checkpoint-sensitive. This pattern is consistent with the training design: action-token CE exposes the action pathway to VLM semantic priors, so tasks dominated by semantic-grounding decisions can benefit from the backbone's strengths in semantic reasoning, while tasks requiring precise low-level control (fine-grained insertion, deformable folding) remain bottlenecked by downstream execution.

\paragraph{Held-out tasks improve with seen tasks.} From \Cref{fig:zeroshot_trend}, the average task progress of unseen tasks ($24.2\% \to 53.6\%$) and seen tasks ($26.1\% \to 50.0\%$) both improve overall despite checkpoint-level fluctuations, with unseen tasks (53.6\%) slightly exceeding seen tasks (50.0\%) at 400k. Because the seen and unseen groups are not difficulty-matched, the trend is more informative than the absolute ordering. The parallel improvement makes it less likely that the model relies only on memorized tasks within the training distribution and instead suggests manipulation capabilities that transfer across scenes and tasks. In the context of large-scale pretraining, however, "unseen" is a relative concept: these tasks have never been collected as identical task data on the current embodiment and use entirely new prop combinations, but open-source data in the pretraining corpus may contain semantically related manipulation experience. The generalization here thus primarily reflects cross-scene, cross-prop skill transfer rather than learning entirely novel skills from scratch. Notably, Rope Tightening (deformable manipulation) reaches task progress 82 at 400k, providing a strong example of transfer in a held-out deformable scenario.

\paragraph{Zero-shot capability has clear limits.} Based on 400k performance, tasks can be categorized into three tiers:

\begin{itemize}
    \item \emph{Zero-shot proficient} (task progress $\geq$ 60\%): Block Sorting (100\%), Fruit Sorting (96\%), Ring Stacking (86\%), Rope Tightening (82\%), etc. These tasks are characterized by clear semantics and moderate precision requirements, where VLM semantic understanding provides a strong scaffold for effective actions.
    \item \emph{Partially proficient} (task progress 40\%--60\%): Switch Pressing (55\%), Number Ordering (54\%), Flower Arranging (51\%), Package Sorting (48.5\%), etc. The model has mastered basic operations (approach, grasp, move) but still falls short on final precise execution. This gap precisely defines the capability delta that fine-tuning needs to bridge.
    \item \emph{Currently beyond zero-shot reach} (task progress $<$ 20\%): Towel Folding (10\%), Table Setting (9\%), Charger Plugging (9\%). These tasks primarily involve deformable objects and fine-grained manipulation. We attribute this to two factors: first, these tasks are inherently far more difficult than rigid pick-and-place, with precision and state perception requirements exceeding what can be achieved without fine-tuning; second, the action patterns required (e.g., pinching fabric edges, folding along trajectories, fine-adjusting alignment for insertion) are relatively isolated in the overall dataset, with low overlap with common task action distributions, making effective cross-task transfer difficult.
\end{itemize}

\paragraph{Capabilities emerge during pretraining.} Multiple tasks show breakthrough progress at specific training stages (see \Cref{sec:appendix_zeroshot} for details): Block Sorting jumps from approximately 50\% to 100\% in the mid-to-late period, Ring Stacking crosses into the fully successful regime by reaching 100\% at 350k before remaining strong at 86\% at 400k, and Fruit Sorting rises from 61\% to 96\%. This "staircase" emergence pattern resembles the emergent abilities observed in large language models: capabilities may first appear as sharp checkpoint-level breakthroughs before stabilizing into the broader pretrained policy. Meanwhile, the overall average task progress at 400k continues to rise (seen 50.0\%, unseen 53.6\%), suggesting that pretraining may not yet be saturated and that further scaling could yield additional gains.

\begin{table}[ht]
\centering
\caption{Highlighted zero-shot tasks at the 400k checkpoint (task progress $\geq$ 60). \emph{Unseen} tasks are bolded; deformable tasks are particularly notable for generalization. Full per-task results in \Cref{sec:appendix_zeroshot}.}
\label{tab:zeroshot_highlight}
\setlength{\tabcolsep}{5pt}
\renewcommand{\arraystretch}{1.12}
\begin{tabularx}{0.98\textwidth}{@{}>{\raggedright\arraybackslash}p{0.25\textwidth}
    >{\raggedright\arraybackslash}p{0.29\textwidth}
    >{\centering\arraybackslash}X
    >{\centering\arraybackslash}p{0.16\textwidth}@{}}
\toprule
\textbf{Task} & \textbf{Category} & \textbf{Seen / Unseen} & \textbf{Task progress} \\
\midrule
Block Sorting        & Semantic understanding & Seen           & \textbf{100\%} \\
Fruit Sorting        & Semantic understanding & Seen           & 96  \\
Ring Stacking        & Rigid manipulation & Seen           & 86  \\
Rope Tightening      & Deformable manipulation & \textbf{Unseen} & 82  \\
Cup Grasping         & Rigid manipulation & Seen           & 64  \\
Bean Pouring         & Deformable manipulation & \textbf{Unseen} & 60  \\
\bottomrule
\end{tabularx}
\end{table}

\subsection{Real-Robot Fine-Tuning Results}\label{sec:finetune}

\subsubsection{Baseline Comparison}\label{sec:baseline}

We benchmark against two pretrained robot foundation models from the two dominant paradigms: \pimodel{}, a vision-language-action (VLA) model, and DreamZero, a world-action model (WAM). All three models (ours, \pimodel{}, and DreamZero) are initialized from their respective official pretrained weights, and are fine-tuned and evaluated on 15 real-robot tasks (10 manipulation + 5 reasoning) under identical fine-tuning data (${\sim}500$ demonstration trajectories per task) and evaluation protocols.

\begin{table}[ht]
\centering
\caption{Real-robot fine-tuning baseline comparison. Values are average task progress per category (max 100). Per-task details in \Cref{sec:appendix_finetune}.}
\label{tab:baseline}
\begin{tabular}{lccc}
\toprule
Model & Manipulation (10) & Reasoning (5) & Overall (15) \\
\midrule
\walloss{} & \textbf{61.1} & \textbf{59.3} & \textbf{60.5} \\
\pimodel{} & 35.0 & 58.9 & 43.0 \\
DreamZero & 33.7 & 32.7 & 33.4 \\
\bottomrule
\end{tabular}
\end{table}

\paragraph{Overall result.} \walloss{} achieves 60.5\% average task progress, outperforming \pimodel{} (43.0\%) by 17.5\% and DreamZero (33.4\%) by 27.1\%, with the highest score on 10 of 15 tasks. \walloss{} now leads on both subsets simultaneously: manipulation (61.1\% vs.\ 35.0\%) and reasoning (59.3\% vs.\ 58.9\%).

\paragraph{Manipulation advantage.} On tasks such as Color Block Sorting (96\% vs.\ 42\%), Ring Stacking (91\% vs.\ 60\%), Drawer Organization (52\% vs.\ 7\%), and Spoon-in-Bowl (80\% vs.\ 43\%), \walloss{} leads \pimodel{} by 30\% or more. These tasks span semantic understanding, medium-precision positioning, and multi-step manipulation. Glasses Rack Placement (66\% vs.\ 87\%) is the only manipulation task where \pimodel{} leads.

\paragraph{Reasoning tasks.} \walloss{} leads \pimodel{} on three of five reasoning tasks: Shape Sorting (68\% vs.\ 63\%), Earphone Sorting (82\% vs.\ 73\%), and Sequential Button Pressing (16\% vs.\ 13\%). These gains drive the reasoning-subset average to 59.3\% vs.\ \pimodel{}'s 58.9\%. Fruit Basket Placement (86\% vs.\ 94\%) and Object Matching (44.5\% vs.\ 51.5\%) remain the two task-level gaps to \pimodel{} on the reasoning subset.

\paragraph{Hard remaining case.} Pencil Case Packing (18.5\%) remains the main low-score manipulation case for \walloss{}, reflecting genuine difficulty in fine-grained bimanual manipulation involving deformable and articulated interactions under the current fine-tuning data budget.

\paragraph{Fine-tuning builds on pretraining.} The strongest post-fine-tuning tasks (Color Block Sorting 96\%, Ring Stacking 91\%) already exhibited high task progress in the \Cref{sec:zeroshot} zero-shot evaluation (100\% and 86\%, respectively), while the tasks with the largest fine-tuning gains (e.g., Drawer Organization jumping from a low zero-shot baseline to 52\%) demonstrate fine-tuning's amplification effect on pretrained base capabilities. This suggests that the capability foundation established during pretraining significantly influences the ceiling of fine-tuning performance.

\subsubsection{Multi-Task Fine-Tuning Scaling}\label{sec:scaling}

\paragraph{Setup.} To evaluate the pretrained model's fine-tuning capability in multi-task settings, we progressively expand the fine-tuning task set from 5 to 10 to 19 tasks, all fine-tuned from the same pretrained checkpoint, analyzing performance trends as task scale grows. The three configurations are:

\begin{itemize}
    \item \textbf{5 tasks}: A basic manipulation task subset selected from \Cref{sec:zeroshot}.
    \item \textbf{10 tasks}: The 5-task set augmented with 5 higher-difficulty or multi-step reasoning tasks.
    \item \textbf{19 tasks}: The 10-task set further augmented with 9 additional manipulation tasks from the broader real-robot evaluation suite. These 9 added tasks differ notably from the original 10 in background environments and embodiment distributions, so this configuration expands not only task count but also scene and object coverage.
\end{itemize}

All experiments use identical training configurations with a uniform 6 epochs, ensuring each task receives the same number of training passes. Overall results are shown in \Cref{fig:scaling}; per-task details, including Towel Folding, are in the appendix. Towel Folding lies outside the scope of this scaling study; see \Cref{sec:appendix_multitask}.

\begin{figure}[t]
    \centering
    \includegraphics[width=\linewidth]{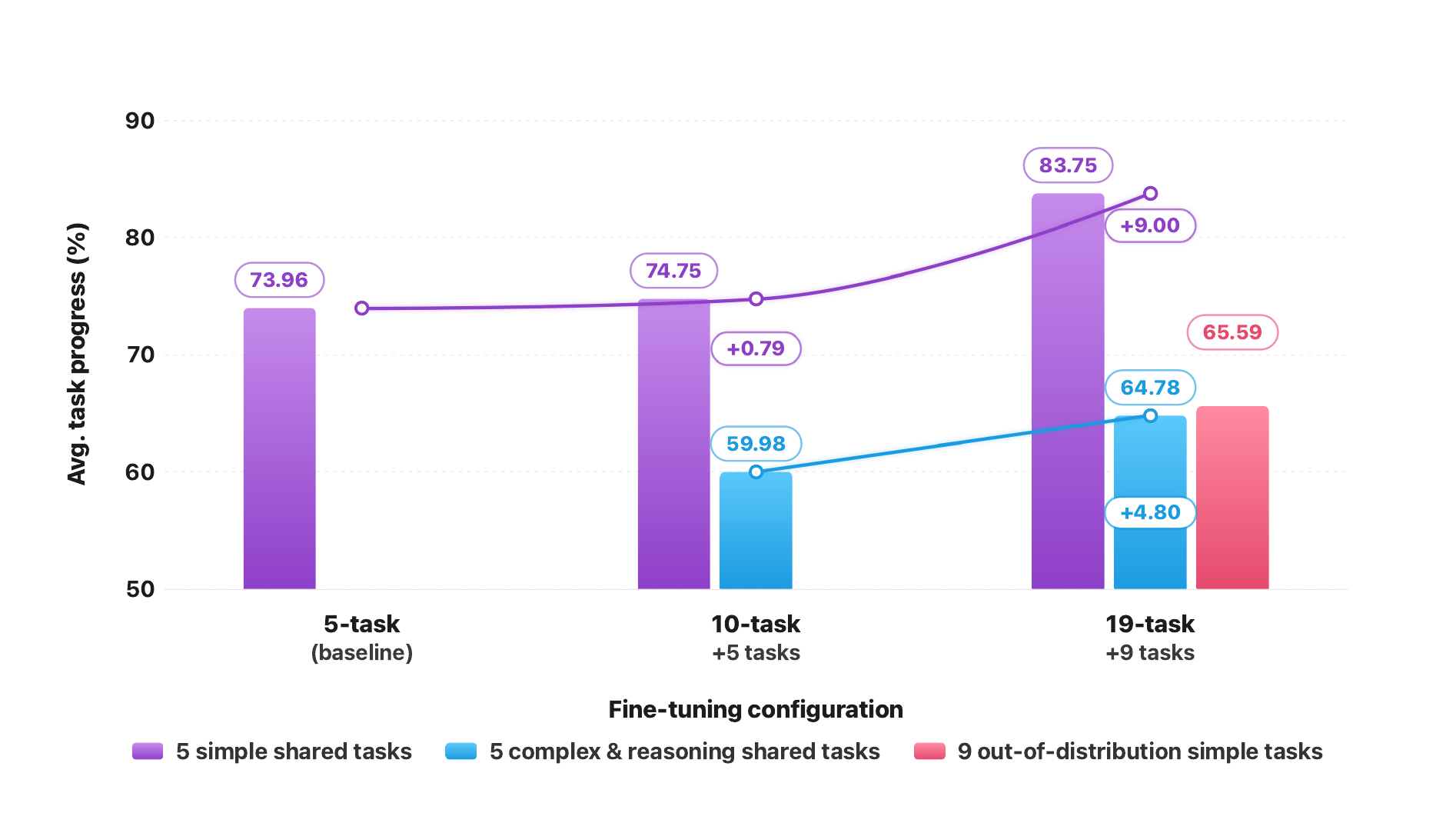}
        \caption{Multi-task fine-tuning scaling results on real-robot tasks. The figure compares models fine-tuned on progressively larger task sets (5, 10, and 19 tasks) and reports average task progress on the shared 5 simple-task subset, the shared 10-task subset, and the 9 newly added out-of-distribution simple tasks. Expanding the fine-tuning set improves performance on shared tasks rather than diluting them: the 5-task subset rises from 73.96\% to 83.75\%, the 10-task subset improves from 59.98\% to 64.78\%, and the added out-of-distribution tasks reach 65.59\% under the 19-task configuration.}
        \label{fig:scaling}

\end{figure}

\paragraph{Task expansion helps shared tasks.} As the fine-tuning task set expands, performance on shared subsets improves monotonically. On the 5 simple shared tasks, scaling 5\,$\to$\,10\,$\to$\,19 tasks raises the average from 73.96\% to 74.75\% to \textbf{83.75\%} (a +9.8\% gain end-to-end). On the 10 shared tasks (5 simple + 5 complex \& reasoning), scaling 10\,$\to$\,19 raises the average from 59.98\% to \textbf{64.78\%} (+4.8\%). Notably, the gains on shared subsets persist after introducing 9 additional tasks with distinctly different backgrounds and embodiment distributions, where the model also reaches 65.59\% average on these out-of-distribution tasks. This shows that under the current model capacity and training setup, task expansion benefits not only adaptation to new tasks but also sustained improvement on shared tasks.

\paragraph{Gains come from broader capability coverage.} We hypothesize that this improvement more likely stems from supplementation at the fine-grained capability level rather than direct whole-task transfer. Although the 9 added tasks differ substantially from the original 10 in overall form, they may still supplement the original training data's coverage gaps in atomic action patterns, language instruction expressions, and state change distributions. For example, basic approach, align, grasp, adjust, and place action primitives may be more thoroughly observed in the new tasks; similarly, different object description styles, object reference forms, and environmental perturbation conditions may expand the model's coverage of the action-language-state combinatorial space. The new tasks' contribution thus more likely manifests as completing reusable intermediate capabilities, improving generalization and robustness on original tasks rather than simply learning a set of independent new task policies.

\subsection{Embodied Multimodal Understanding}\label{sec:multimodal}

To evaluate the impact of co-training on multimodal understanding capabilities, we select 5 representative benchmarks for evaluation (detailed scores in \Cref{sec:appendix_multimodal}), divided into two groups: general visual question answering (RealWorld VQA~\citep{realworldqa}, ERQA~\citep{gemini_robotics}) and embodied understanding directly relevant to robot execution (EO-Bench~\citep{eo1_2025} scene understanding, Embodied Grounding target localization, Where2Place~\citep{RoboPoint} placement reasoning), with the VLM backbone Qwen2.5-VL-3B~\citep{Qwen2.5-VL} as baseline. Embodied Grounding is an internally constructed benchmark whose samples are drawn and annotated from robot action trajectories.

\begin{figure}[!tbp]
    \centering
    \includegraphics[width=\linewidth]{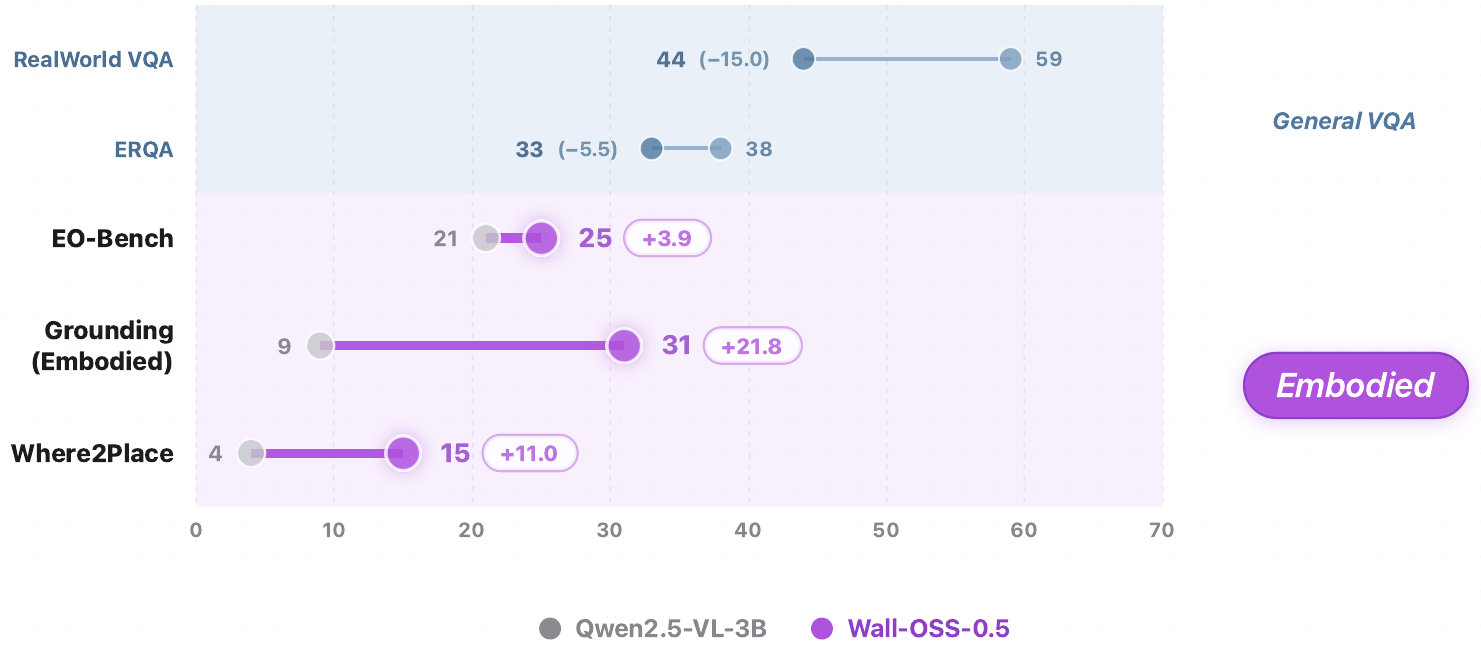}
    \caption{Multimodal capability changes from co-training, measured relative to the backbone Qwen2.5-VL-3B.}
    \label{fig:vlm_benchmarks}
\end{figure}

\Cref{fig:vlm_benchmarks} shows the score changes of \walloss{} relative to the backbone on each benchmark. The results suggest a \emph{specialization} effect from co-training: performance shifts away from open-domain VQA and toward embodied perception signals more relevant to robot execution.

\paragraph{Significant embodied understanding gains.} The most pronounced improvement is in robot manipulation-oriented target localization (Embodied Grounding, +21.8), which requires the model to localize manipulation targets in the robot's ego-centric view (e.g., "find the red cup on the table"). This differs from open-domain grounding settings: the backbone has a low baseline (9.0) on our embodied manipulation grounding benchmark, which lies outside its original pretraining focus. Placement reasoning Where2Place (+11.0) and embodied scene understanding EO-Bench (+3.9) also show clear gains. These improvements correspond closely to the core perception needs in the robot execution pipeline: "where to look, where to point, where to place". In pilot pretraining runs, we observed that the strong action-token prediction objective can substantially reduce multimodal scores on common robot-observation distributions when embodied bridge data is insufficient. The 12 million bridge samples described in \Cref{sec:multimodal_data} were therefore introduced not only as extra VQA data, but as robot-view grounding and spatial-decision supervision that counterbalances the specialization pressure from action-token CE.

\paragraph{General VQA regression under specialization.} General visual question answering regresses (RealWorld VQA $-15.0$, ERQA $-5.5$), consistent with the specialization pressure introduced by VLA co-training. For a VLA model whose primary objective is action execution, this trade-off is reasonable: the model gives up part of its open-domain VQA performance while improving embodied spatial decision signals, rather than competing with dedicated VLMs on general image understanding.

\begin{figure}[!t]
    \centering
    \includegraphics[width=0.9\linewidth,height=0.7\textheight,keepaspectratio]{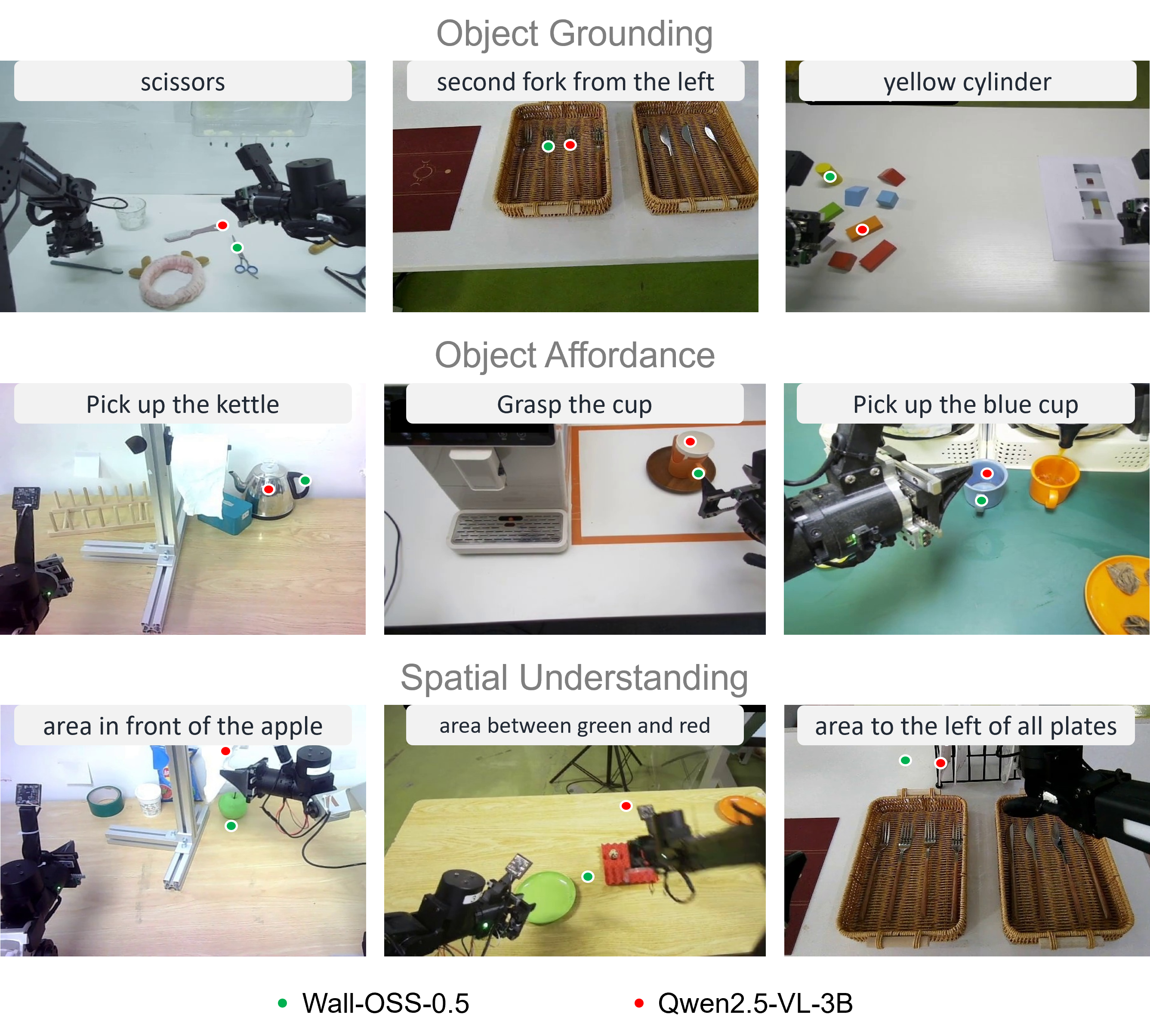}
    \caption{Multimodal understanding visualization comparison in embodied scenes (\walloss{} vs.\ Qwen2.5-VL-3B).}
    \label{fig:grounding_multimodal}
\end{figure}

\paragraph{Visualization examples.} \Cref{fig:grounding_multimodal} shows grounding and spatial reasoning comparisons between \walloss{} and the backbone Qwen2.5-VL-3B in embodied scenes. In the robot's ego-centric view, \walloss{} more accurately localizes manipulation targets and selects placement positions that align with actionable regions, while the backbone is more likely to drift toward visually salient but less task-relevant areas. These examples visually illustrate the embodied perception enhancement from co-training.

%% file: sec/5_ablation.tex
\section{Ablation Studies}\label{sec:ablation}

\subsection{Evaluation of Co-Training Strategies}\label{sec:ablation_strategy}

We compare four training strategies trained from scratch (with only the VLM backbone initialized from pretraining): co-training, stop-gradient, stop-gradient to co-training, and flow-only. All experiments run for 70k training steps on 5 real-robot ablation tasks, under identical configurations. These 5 tasks are a subset selected for tractable from-scratch training; their absolute numbers are not directly comparable to the 17-task pretrained evaluation in \Cref{sec:zeroshot}. This experiment only supports relative comparison between the four strategies. The strategies differ in whether flow-matching gradients reach the VLM backbone and whether action-token cross-entropy is present:

\begin{itemize}[leftmargin=*,itemsep=2pt]
    \item \emph{Co-training} (full gradient bridge): action token cross-entropy, flow matching, and multimodal cross-entropy are jointly optimized, with all gradients backpropagating to the VLM backbone.
    \item \emph{Stop-gradient} (flow gradient blocked from the backbone): the Action Expert is still optimized by flow matching loss, while the backbone is updated only by action token and multimodal cross-entropy. This matches the knowledge-isolation strategy proposed by~\citep{ki2024}.
    \item \emph{Stop-gradient to co-training} (two-stage warm-up): trained under stop-gradient first, then switched to co-training through by removing the gradient block.
    \item \emph{Flow-only} (action token cross-entropy removed): only flow matching and multimodal cross-entropy are optimized; the VLM backbone remains trainable, but receives no signal from action tokens.
\end{itemize}

\begin{figure}[t]
    \centering
    \begin{subfigure}[t]{0.49\linewidth}
        \includegraphics[width=\linewidth]{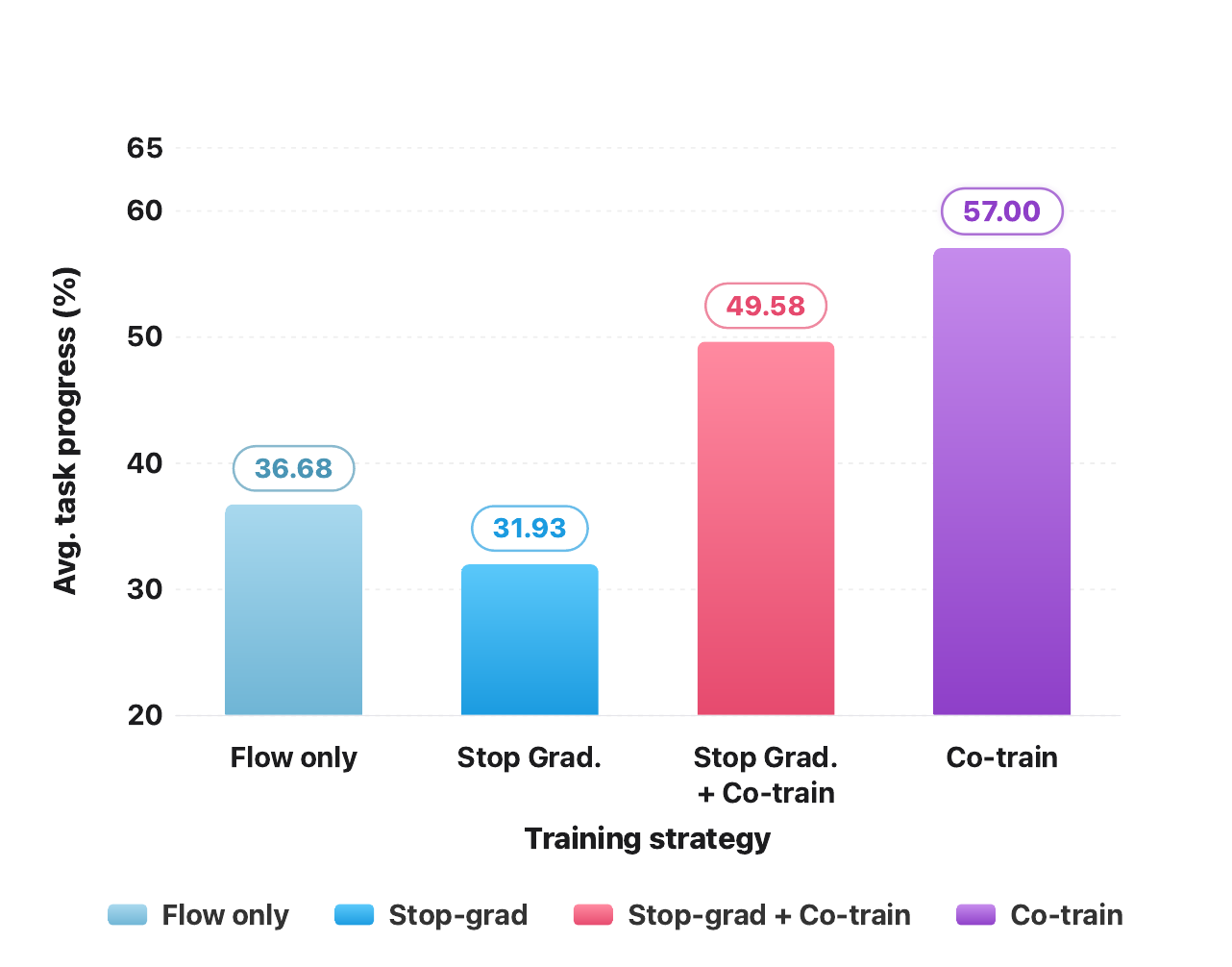}
        \caption{From-scratch results on 5 real-robot ablation tasks.}
        \label{fig:ablation_strategy}
    \end{subfigure}
    \hspace{-0.02\linewidth}
    \begin{subfigure}[t]{0.5\linewidth}
        \includegraphics[width=\linewidth]{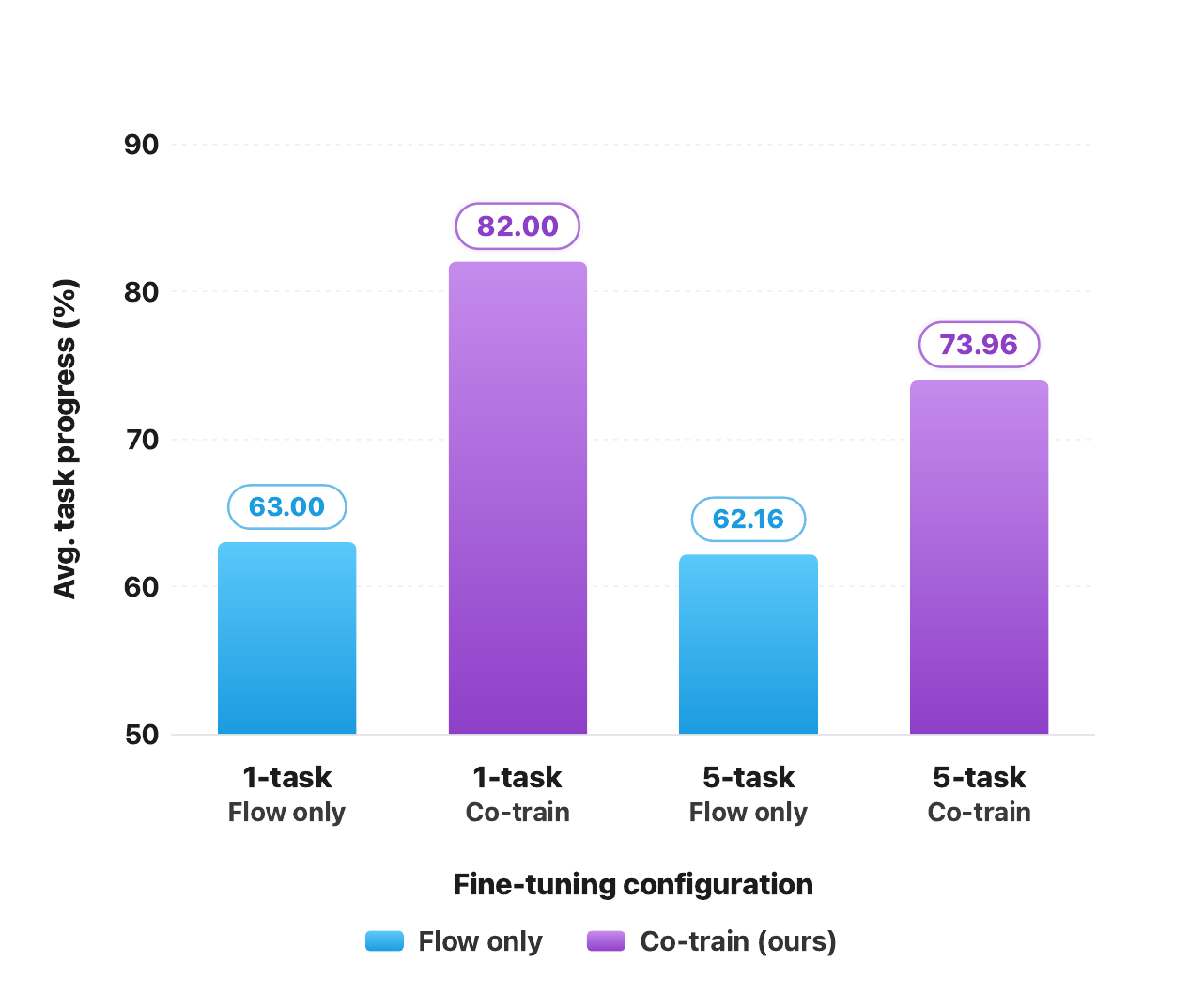}
        \captionsetup{margin={0.06\linewidth,0pt}}
        \caption{Fine-tuning stage results.}
        \label{fig:ablation_finetune}
    \end{subfigure}
    \caption{Training strategy comparison. (a) Co-training achieves the best from-scratch performance (70k steps). (b) Co-training advantages persist during fine-tuning.}
    \label{fig:ablation_strategy_combined}
\end{figure}

As shown in \Cref{fig:ablation_strategy}, co-training reaches an average task progress of 57.0\%, outperforming flow-only (36.6\%), stop-gradient (31.9\%), and stop-gradient to co-training (49.6\%) on 5 ablation tasks. This supports the gradient-bridged co-training claim in \Cref{sec:intro}: removing any of the three signals (flow alone, no bridge to backbone, or two-stage substitution) degrades real-robot performance by 7.4\,--\,25.1 percentage points. Across all four strategies, the VQA scores stay tightly clustered, with stop-gradient marginally ahead of the other three.

Stop-gradient, the strategy marginally ahead on VQA, achieves the lowest score among all action tasks, and this gap is consistent across all tasks. During training, stop-gradient exhibits slower flow-loss convergence and a higher final loss value, indicating Action Expert underfitting. Co-training's flow loss, in contrast, converges faster than both flow-only and stop-gradient, with a final value lower than stop-gradient and comparable to flow-only. Based on these observations, co-training is adopted as the training strategy for the pretraining stage.

This finding extends beyond pretraining: during fine-tuning, discrete action tokens likewise provide more efficient adaptation signals to the VLM backbone, accelerating learning on new tasks, as shown in \Cref{fig:ablation_finetune}. Co-training with discrete action tokens and multimodal dataset is therefore also adopted for the fine-tuning stage.

\subsection{Effect on Action-Space Loss}\label{sec:ablation_flowloss}

To compare the action-space loss (\Cref{sec:action_space}) with the traditional flow-matching loss in velocity fields, we carry out controlled experiments in the LIBERO~\citep{liu2023libero} simulation environment. Both conditions use the same model architecture as the pretraining backbone, with weights initialized from Qwen2.5-VL and the Action Expert trained from scratch.  Training uses a global batch size of 128, with all remaining hyperparameters held identical across the two conditions.

\begin{figure}[H]
\centering
\begin{minipage}[t]{0.55\linewidth}
\vspace{0pt}
\centering
\includegraphics[width=\linewidth]{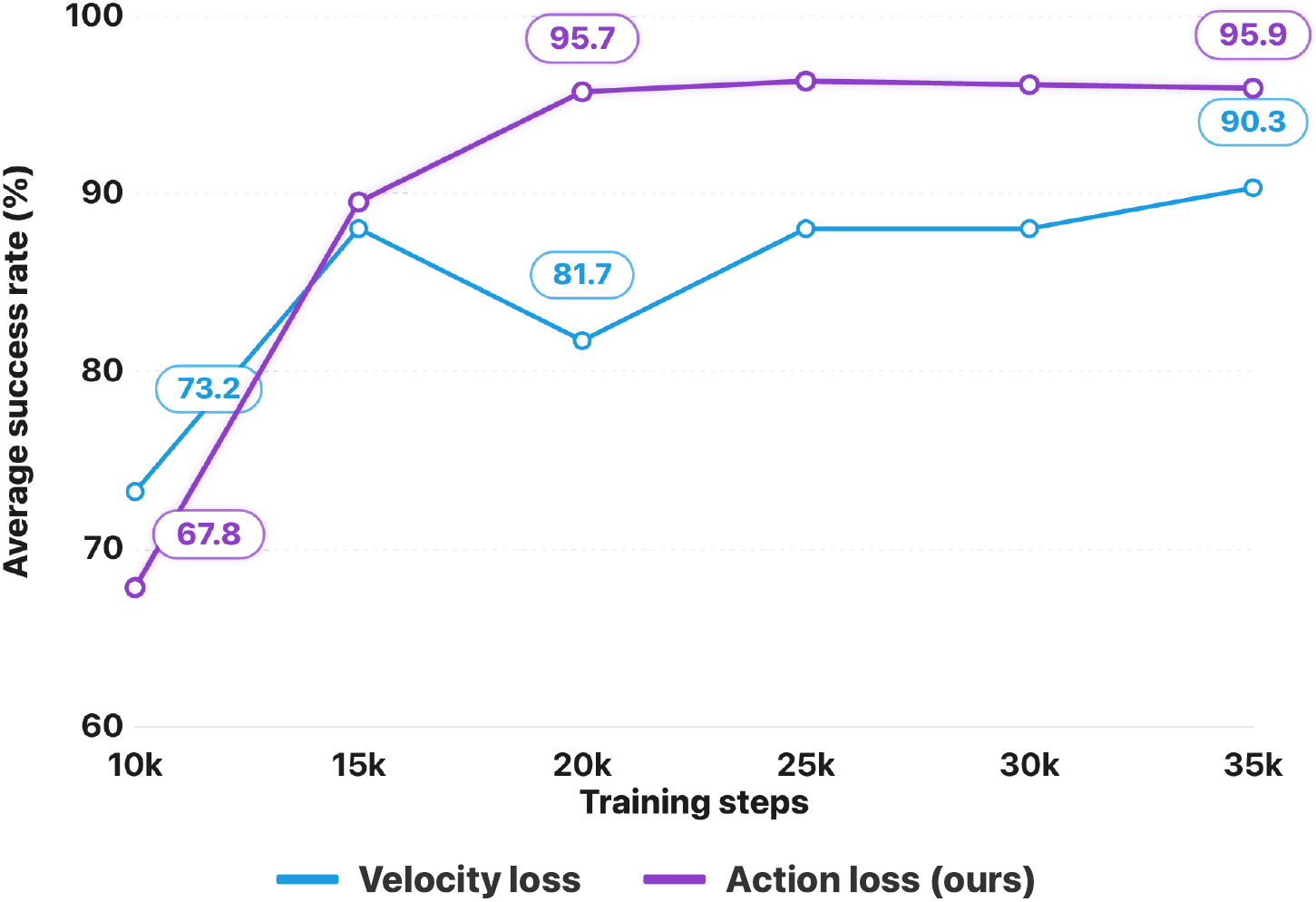}
\captionof{figure}{
action-space loss vs.\ velocity-space loss on LIBERO.
}
\label{fig:ablation_flowloss}
\end{minipage}\hfill
\begin{minipage}[t]{0.43\linewidth}
\vspace{0pt}

\smallskip
\textbf{Training stability.} The velocity-space loss exhibits significant performance fluctuations at 20k, while the action-space loss consistently maintains the 92.5\%--96.5\% range after 20k steps. The only exception occurs in early training (10k steps), where the velocity-space loss slightly outperforms the action-space loss. This behavior is consistent with the design rationale of Action-Space Supervision: since the action-space loss is equivalent to $(1-\tau)^2$-reweighting the velocity-space loss, gradient signals in the low-noise regime are relatively sparse during early training. Once the model has acquired the low-frequency trajectory structure through high-noise supervision, performance rapidly improves.
\end{minipage}
\end{figure}

\smallskip
\textbf{Performance.} As shown in \Cref{fig:ablation_flowloss}, the action-space loss variant achieves 96.5\% peak average success rate at 25k steps, exceeding the velocity-space loss peak by 6.2\%.
For the convergence speed, the action-space loss reaches 95.8\% average success rate at only 20k training steps, while the velocity-space loss fails to surpass 90.3\% after the full 35k steps.

These experimental results support defining the loss function in the action space and align with the spectral characteristics of robot action signals: useful information concentrates in low-frequency components, and the high-noise regime constitutes the dominant quality bottleneck.

\subsection{Effect of the Vision-Aligned RVQ Action Tokenizer}\label{sec:ablation_tokenizer}

\begin{figure}[ht]
    \centering
    \includegraphics[width=\linewidth]{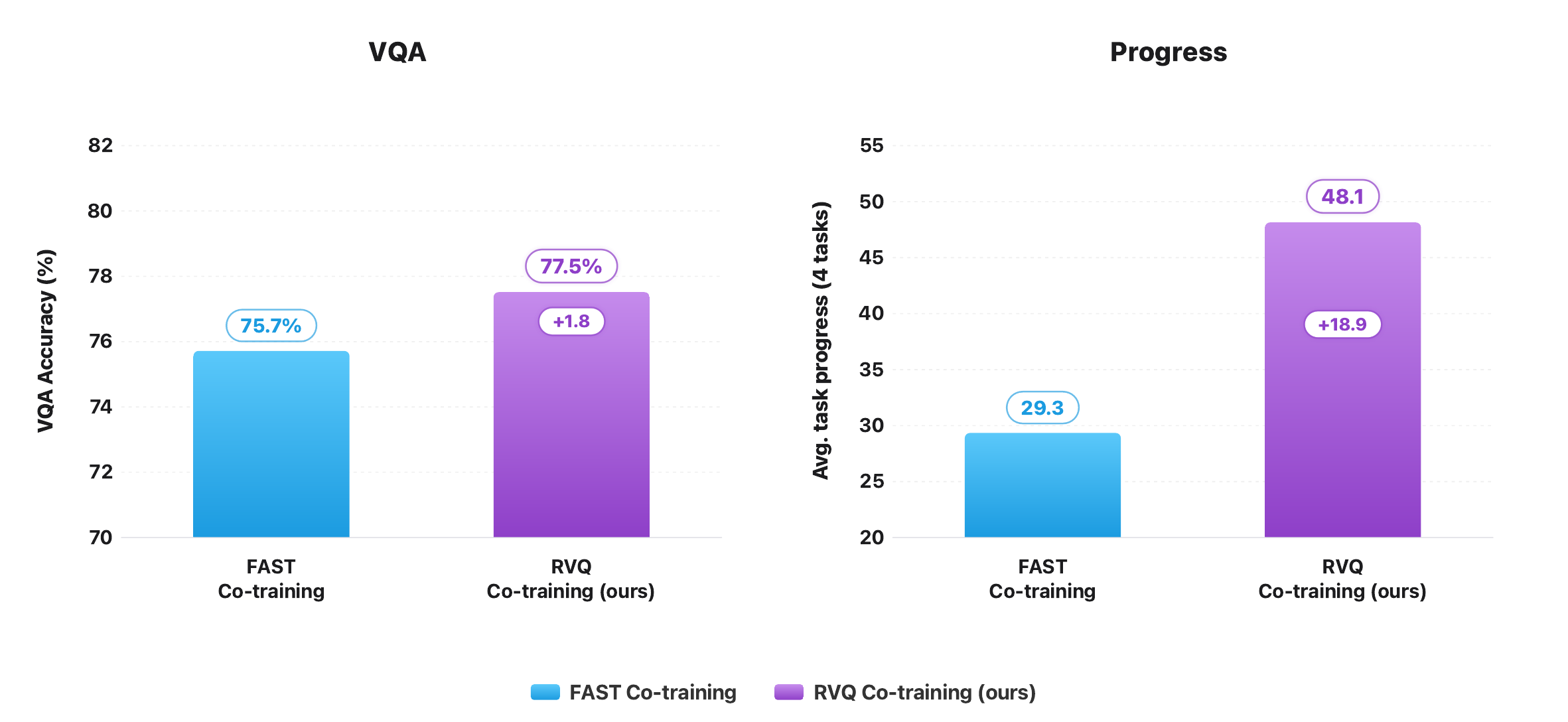}
    \caption{Vision-Aligned RVQ Action Tokenizer vs.\ FAST tokenizer under identical co-training settings. Action task values are average task progress over 4 real-robot tasks (max 100), with RVQ leading on every task individually.}
    \label{fig:ablation_tokenizer}
\end{figure}

To validate the effectiveness of the proposed Vision-Aligned RVQ Action Tokenizer relative to the FAST tokenizer~\citep{fast}, we conduct a controlled experiment that replaces only the tokenizer under identical co-training settings, training fully from the VLM backbone weights. Evaluation covers four real-robot tasks and one VQA task; real-robot evaluation uses the continuous actions generated by the flow-matching pathway under co-training.

\paragraph{Multimodal understanding.} As shown on the left of \Cref{fig:ablation_tokenizer}, our proposed RVQ Action Tokenizer not only preserves VQA performance but improves accuracy from 75.7\% to 77.5\%. This indicates that the auxiliary multimodal objectives used to train the tokenizer, visual-action alignment and next-frame prediction (\Cref{sec:vq_tokenizer}), shape token representations that also contribute to the VLM's overall visual-semantic understanding.

\paragraph{Task Performance.} As shown on the right of \Cref{fig:ablation_tokenizer}, our proposed RVQ Action Tokenizer improves average task progress from 29.3\% to 48.1\%. Notably, real-robot evaluation uses continuous actions generated by flow matching rather than direct decoding of discrete tokens,  so the tokenizer's gains are not confined to the discrete pathway: higher-quality discrete representations during co-training also enhance the quality of continuous action generation.

%% file: sec/6_related.tex
\section{Related Work}\label{sec:related}

\paragraph{Action representations.} A primary approach for adapting pretrained vision-language models (VLMs) into Vision-Language-Action (VLA) systems involves directly modeling continuous action distributions, typically through diffusion (e.g., Diffusion Policy~\citep{diffusion_policy}) or flow matching~\citep{lipman2022flow}. In contrast, \walloss{} focuses on discrete and latent action representations. Early efforts such as RT-2~\citep{rt2} and OpenVLA~\citep{openvla} discretize robot actions into text-like tokens, demonstrating that VLM semantic knowledge can transfer to robot control while exposing the precision limits of naive per-dimension discretization. Although FAST~\citep{fast} enhances token efficiency via DCT, it remains a rule-based compressor with limited semantic capacity. Other reconstruction-driven approaches, including VQ-BeT~\citep{lee2024behavior}, FasTer~\citep{liu2025faster}, and OAT~\citep{liu2026orderedactiontokenization}, likewise confine action tokenization to geometric signal compression. These methods formulate tokenization exclusively from an action-centric perspective. By overlooking contextual information such as environmental states and task intents, they focus on minimizing numerical trajectory reconstruction errors rather than capturing abstract behavioral meanings.Recent advancements favor semantic and context-aware representations, branching into two main strategies. The first incorporates generative semantic tokenization, where methods like ActionCodeC~\citep{dong2026actioncodec}, LipVQ-VAE~\citep{vuong2025action}, LAPA~\citep{ye2024lap}, and UniAct~\citep{zheng2025universal} project trajectories into discrete latent action tokens to capture underlying behavioral dynamics. The second strategy emphasizes information preservation through contrastive learning; frameworks such as DecisionNCE~\citep{li2024decisionnce} and CLAP~\citep{zhang2026clapcontrastivelatentaction} utilize cross-modal contrastive objectives to map actions into a semantically structured space aligned with video or language contexts. Furthermore, UniT~\citep{unit2026} introduces a unified physical language for human-to-humanoid policies, underscoring the efficacy of such contrastive latent tokens in bridging human demonstration and humanoid world modeling. The closest comparison is \pimodel{}~\citep{pi05}, which co-trains a FAST-based~\cite{fast} autoregressive pathway with a flow-matching pathway. \walloss{} differs in three respects: it employs a learned Vision-Aligned RVQ Action Tokenizer, analyzes the discrete pathway as a gradient bridge, and introduces Action-Space Supervision for flow matching. A separate line of work, including GR-2~\citep{gr2}, DreamZero~\citep{dreamzero}, and LingBot-VA~\citep{lingbot-va2026}, learns world dynamics through video generation before transferring to action prediction.

\paragraph{Architectures for action-ready VLMs.} A primary design challenge for VLAs is integrating action generation without overwriting the semantic knowledge stored in the VLM backbone. Existing solutions span multiple paradigms. The earliest generalist controllers train a single transformer from scratch over multi-task robot data, bypassing VLM transfer entirely. Notable examples include RT-1~\citep{rt1}, Gato~\citep{gato}, Octo~\citep{octo}, and the bimanual diffusion model RDT-1B~\citep{liu2024rdt}. A second family reuses a pretrained VLM and exposes actions as additional discrete tokens consumed by the same shared transformer. This category includes RT-2~\citep{rt2}, OpenVLA~\citep{openvla}, RoboFlamingo~\citep{li2023vision}, TinyVLA~\citep{wen2025tinyvla}, X-VLA~\citep{xvla}, SpatialVLA~\citep{qu2025spatialvla}, and 3D-VLA~\citep{zhen20243d}, which preserves the language interface but ties action precision to the tokenization scheme. A third family keeps the VLM trunk and attaches a continuous-action head. For example, the \pizero{}~\citep{pi0,pi05,intelligence2026pi} family adds a flow-matching action expert that cross-attends to a largely frozen backbone, decoupling reasoning capacity from low-level control precision. A fourth family enforces a hierarchical or dual-system split between reasoning and execution. For instance, Gemini Robotics~\citep{gemini_robotics}, GR00T-N1~\citep{gr00tn1_2025}, and Galaxea G0~\citep{galaxea_openworld} pair a high-capacity VLM planner with a lighter controller, whereas Hi Robot~\citep{shi2025hi}, Embodied-CoT~\citep{zawalski2024embodiedcot}, CoT-VLA~\citep{zhao2025cot}, OneTwoVLA~\citep{lin2025onetwovla}, and ChatVLA~\citep{zhou2025chatvla} introduce explicit reasoning steps before action emission. A fifth family replaces the VLM trunk with a video-generative world model, treating actions as the consequence of a predicted next observation, as seen in GR-1~\citep{gr1}, GR-2~\citep{gr2}, Cosmos Policy~\citep{kim2026cosmos}, and mimic-video~\citep{pai2025mimic}. A sixth family augments the policy with explicit memory or temporal context modules, such as MemoryVLA~\citep{shi2025memoryvla}, ContextVLA~\citep{jang2025contextvla}, CronusVLA~\citep{li2025cronusvla}, TraceVLA~\citep{zheng2025tracevla}, MEMER~\citep{sridhar2025memer}, and SAM2Act~\citep{fang2025sam2act}, to extend the effective temporal horizon. Closest to our work, CogACT~\citep{cogact} splits cognition and action into independent modules, and HPT~\citep{hpt} uses embodiment-specific stems coupled to a shared trunk. \walloss{} extends this co-design by adopting a Mixture-of-Transformers architecture with a VL Expert and an Action Expert at each Transformer layer. The two experts interact through shared attention while retaining separate parameters, shifting the role of expert partitioning from generic knowledge sharding~\citep{moe_mixtral,deepseek_moe} to modality and function routing.

\paragraph{Robot data and embodied grounding.} Open X-Embodiment~\citep{oxe}, DROID~\citep{droid}, RoboMIND v1~\citep{robomind}, RoboMIND v2~\citep{robomind2}, RoboCOIN ~\citep{robocoin}and AgiBot World~\citep{agibot_world} have continued to expand the scale and diversity of available robot data. Our pretraining mixture is comparable in scale, with over one million trajectories per epoch across more than 20 embodiments, but is organized around deployment rather than scale alone. We apply task-level power sampling to mitigate long-tail imbalance and supplement the mixture with large-scale embodied bridge samples synthesized from action corpora, linking visual understanding to executable action semantics rather than relying on generic VQA data alone~\citep{rt2,pi05}.

%% file: sec/7_discussion.tex
\section{Discussion and Limitations}\label{sec:discussion}

This report advances an operational claim: sufficiently large-scale VLA pretraining can already produce measurable real-world robotic behavior, while the resulting checkpoint simultaneously serves as a substantially stronger prior for downstream adaptation.
We conclude by discussing the broader design implications, current limitations, and promising future directions.

\paragraph{Design implications.}
The gradient-bridge mechanism suggests that discrete action tokens remain valuable even when deployment ultimately relies on continuous actions. Their primary role emerges during pretraining: action-token cross-entropy provides a strong, VLM-native supervisory channel that directly shapes the backbone toward controllable representations. In contrast, continuous-only training exposes the backbone primarily to a relatively weak residual flow-matching signal, while stop-gradient formulations decouple representation learning from continuous control optimization, preventing the two objectives from fully reinforcing one another.

This perspective also unifies two seemingly independent design choices. MoT routing (\Cref{sec:overall_arch}) allocates dedicated parameter capacity to continuous-action computation without severing gradients flowing into the VLM backbone. Meanwhile, embodied bridge data (\Cref{sec:multimodal_data}) makes the multimodal anchor more action-aware, enabling the model to preserve grounded understanding in environments closer to those encountered during physical execution. Together, both choices pursue the same objective: transforming a pretrained multimodal model into an executable robotic policy without collapsing it into a narrow task-specific controller.

\paragraph{Limitations.}
Several limitations remain. First, the gradient-bridge dynamics have thus far only been validated with a 3B VLM backbone; scaling to larger VLM backbones may substantially alter the relative geometry and interaction strength among the three training signals. Second, the current model operates on single-frame image inputs, which likely constrains zero-shot performance on long-horizon tasks requiring temporal memory and persistent state tracking. Third, the Vision-Aligned RVQ Action Tokenizer and the associated training pipeline are currently tied to a fixed 26-dimensional action representation, limiting direct applicability to dexterous hands and other high-DoF embodiments that require richer action interfaces. Finally, task evaluation still depends on manually designed scoring rubrics (\Cref{sec:scoring_rubrics}), and the present real-robot benchmark does not yet cover multi-robot collaboration, long-duration deployment, or broader open-world interaction settings.

\paragraph{Future Work.}
Future work will scale the framework to larger VLM backbones, incorporate temporal observations and hierarchical planning for long-horizon tasks, and explore more general action representations capable of supporting diverse robot morphologies. As an open-source project, we will continue releasing model weights, training code, and evaluation tools to facilitate future research and reproducibility.

%% file: sec/contributors.tex
\section{Contributors}\label{sec:contributors}
\walloss{} is a collaborative effort of the X Square Robot team. The full contributor list is given below; $*$ denotes core contributors, $\dagger$ denotes the project lead, and $\ddagger$ denotes the corresponding author.

Ryan Yu$^{*}$, Pushi Zhang$^{*}$, Starrick Liu$^{*}$, Brae Liu$^{*}$, Miracle Kang$^{*}$, Shalfun Li$^{*}$, Lights Shi, Ellie Ma, Ping Yang, Chris Pan, Jerry Chen, Dongxiu Liu, Rain Sun, Miles Guo, Byron Zhang, Hugo Zhou, Zach Xu, Vincent Chen, Harrison Huang, James Wang, Dance Kuzi, Andy Zhai, Hang Su, Roy Gan, Lucy Liang$^{\dagger}$, Hao Wang$^{\ddagger}$, Qian Wang.

%% file: sec/appendix.tex
\section{Appendix}\label{sec:appendix}
\subsection{Pretrained Model Zero-Shot Evaluation Per-Task Results}\label{sec:appendix_zeroshot}

The following tables list per-task task progress for all seen tasks in Table~\ref{tab:zeroshot_seen} and unseen tasks in Table~\ref{tab:zeroshot_unseen} across milestone checkpoints from \cref{sec:experiments}. Each task is evaluated over 10 trajectories, scored according to predefined scoring rubrics (maximum 100).
Category abbreviations (\textbf{Cat.}): \textbf{Sem.}\,=\,semantic understanding, \textbf{Rigid}\,=\,rigid-object, \textbf{Def.}\,=\,deformable-object, \textbf{Fine}\,=\,fine-grained, \textbf{Long}\,=\,long-horizon multi-step manipulation.

\begin{table}[H]
    \centering
    \caption{Seen tasks zero-shot per-task progress. Bold indicates the highest task progress across checkpoints for each task.}
    \label{tab:zeroshot_seen}
    \small
    \begin{tabularx}{\textwidth}{@{}>{\raggedright\arraybackslash}Xlcccccc@{}}
        \toprule
        Task & Cat. & 50k & 100k & 200k & 300k & 350k & 400k \\
        \midrule
        Block Sorting & Sem. & 46 & 85 & 87.5 & 51.5 & 96 & \textbf{100} \\
        Fruit Sorting & Sem. & 41 & 19 & 81 & 61 & 61 & \textbf{96} \\
        Number Ordering & Sem. & 45 & \textbf{65} & 32 & 60 & 56 & 54 \\
        Switch Pressing & Sem. & 7 & 18 & 46 & 30 & 38 & \textbf{55} \\
        Ring Stacking & Rigid & 18 & 18 & 51 & 73 & \textbf{100} & 86 \\
        Cup Grasping & Rigid & 48 & \textbf{70} & 63 & 58 & 56 & 64 \\
        Towel Folding & Def. & 9 & 8 & \textbf{11.5} & 10 & 10 & 10 \\
        Table Setting & Def. & 6 & \textbf{15} & 8 & 10 & 10 & 9 \\
        Paper Shredding & Def. & \textbf{28} & 10 & 23 & 12 & 18 & 18 \\
        Charger Plugging & Fine & 5 & 2 & 2 & 6 & 6 & \textbf{9} \\
        Flower Arranging & Long & 30 & 34.5 & 37.5 & 54.5 & \textbf{57} & 51 \\
        Package Sorting & Long & 29.75 & 36.25 & 39.25 & 58.5 & \textbf{68.75} & 48.5 \\
        \midrule
        \textbf{Seen avg.} & & 26.1 & 31.7 & 40.1 & 40.4 & 48.1 & \textbf{50.0} \\
        \bottomrule
    \end{tabularx}
\end{table}

\begin{table}[H]
    \centering
    \caption{Unseen tasks zero-shot per-task progress. Bold indicates the highest task progress across checkpoints for each task.}
    \label{tab:zeroshot_unseen}
    \small
    \begin{tabularx}{\textwidth}{@{}>{\raggedright\arraybackslash}Xlcccccc@{}}
        \toprule
        Task & Cat. & 50k & 100k & 200k & 300k & 350k & 400k \\
        \midrule
        Toy Basket Plcmt. & Sem. & 45 & 71 & 66 & 60 & \textbf{72} & 58 \\
        Rope Tightening & Def. & 26 & 30 & 54 & 60 & 62 & \textbf{82} \\
        Bean Pouring & Def. & 12 & 50 & 19 & 13 & 50 & \textbf{60} \\
        Table Wiping & Rigid & \textbf{38} & 36 & 32 & 26 & 28 & 38 \\
        Pot Lid Covering & Rigid & 0 & 18 & 23 & 15 & 26 & \textbf{30} \\
        \midrule
        \textbf{Unseen avg.} & & 24.2 & 41.0 & 38.8 & 34.8 & 47.6 & \textbf{53.6} \\
        \bottomrule
    \end{tabularx}
\end{table}

\subsection{Multimodal Understanding Evaluation Detailed Results}\label{sec:appendix_multimodal}

Table~\ref{tab:multimodal_detail} lists detailed scores and task descriptions for each benchmark in \cref{sec:experiments}.

\begin{table}[H]
    \centering
    \caption{Multimodal understanding benchmark detailed scores. \emph{Embodied Grounding} is an internally constructed benchmark; evaluation samples are directly sampled and annotated from robot action dataset trajectories.}
    \label{tab:multimodal_detail}
    \small
    \begin{tabularx}{\textwidth}{@{}ll>{\raggedright\arraybackslash}Xccc@{}}
        \toprule
        Dimension & Benchmark & Task Description & Qwen2.5-VL-3B & \walloss{} & Change \\
        \midrule
        General VQA & RealWorld VQA & Open-ended VQA on real-world scene images & 59.2\% & 44.2\% & $-$15.0 \\
        General VQA & ERQA & Entity recognition and relational QA for embodied scenes & 38.3\% & 32.8\% & $-$5.5 \\
        Embodied scene & EO-Bench & Comprehensive visual understanding in embodied manipulation scenes & 20.8\% & 24.7\% & +3.9 \\
        Embodied grounding & Emb.\ Grounding* & Localizing target objects via point coordinates in robot ego-centric images & 9.0\% & 30.8\% & +21.8 \\
        Placement reasoning & Where2Place & Predicting reasonable placement position coordinates & 4.0\% & 15.0\% & +11.0 \\
        \bottomrule
    \end{tabularx}
\end{table}

\subsection{Fine-Tuning Evaluation Per-Task Detailed Results}\label{sec:appendix_finetune}

\begin{table}[H]
    \centering
    \caption{Fine-tuning evaluation per-task results. Values are task progress (max 100); bold indicates row maximum. All three models use their respective official pretrained weights, fine-tuned and evaluated under identical data (${\sim}500$ trajectories per task) and protocols.}
    \label{tab:finetune_detail}
    \small
    \begin{tabularx}{\textwidth}{@{}>{\raggedright\arraybackslash}Xlccc@{}}
        \toprule
        Task & Cat. & \walloss{} & \pimodel{} & Dream. \\
        \midrule
        Color Block Sort. & Manip. & \textbf{96} & 42 & 27 \\
        Ring Stacking & Manip. & \textbf{91} & 60 & 27 \\
        Spoon-in-Bowl & Manip. & \textbf{80} & 43 & 54 \\
        Obj.-to-Basket & Manip. & 74.8 & 37 & \textbf{97.8} \\
        Glasses Rack & Manip. & 66 & \textbf{87} & 37 \\
        Cup Triangle & Manip. & \textbf{58} & 18 & 25 \\
        Drawer Org. & Manip. & \textbf{52} & 7 & 7 \\
        Power Cord & Manip. & \textbf{50} & 21 & 24 \\
        Water Pouring & Manip. & \textbf{25} & 19 & 12 \\
        Pencil Case & Manip. & 18.5 & 16 & \textbf{26} \\
        Fruit Basket & Reas. & 86 & \textbf{94} & 45 \\
        Earphone Sort. & Reas. & \textbf{82} & 73 & 66 \\
        Object Matching & Reas. & 44.5 & \textbf{51.5} & 13.5 \\
        Shape Sorting & Reas. & \textbf{68} & 63 & 36 \\
        Seq.\ Button & Reas. & \textbf{16} & 13 & 3 \\
        \bottomrule
    \end{tabularx}
\end{table}

\newpage
\subsection{Multi-Task Fine-Tuning Per-Task Detailed Results}\label{sec:appendix_multitask}

{\small
\begin{xltabular}{\textwidth}{@{}>{\raggedright\arraybackslash}Xlcccc@{}}
\caption{Multi-task fine-tuning per-task detailed results. Values are task progress (max 100). Dashes indicate the task was not included in that configuration.}
\label{tab:multitask_detail} \\
\toprule
Task & Cat. & 19-task & 10-task & 5-task & 1-task \\
\midrule
\endfirsthead
\toprule
Task & Cat. & 19-task & 10-task & 5-task & 1-task \\
\midrule
\endhead
\midrule \multicolumn{6}{r}{\emph{Continued on next page}} \\
\endfoot
\bottomrule
\endlastfoot
Cup Grasping & Rigid & 81 & 91 & 82 & 82 \\
Package Sorting & Long & 82.75 & 50.75 & 50.8 & -- \\
Block Sorting & Sem. & 100 & 100 & 100 & -- \\
Switch Pressing & Sem. & 76 & 53 & 70 & -- \\
Flower Arranging & Long & 79 & 79 & 67 & -- \\
Table Setting & Def. & 18 & 18 & -- & -- \\
Charger Plugging & Fine & 35 & 24 & -- & -- \\
Fruit Sorting & Sem. & 90 & 91 & -- & -- \\
Number Ordering & Sem. & 59 & 66 & -- & -- \\
Paper Shredding & Def. & 27 & 27 & -- & -- \\
Cup Triangle Stack. & Long & 60 & -- & -- & -- \\
Spoon-in-Bowl & Rigid & 98 & -- & -- & -- \\
Glasses Rack & Rigid & 65 & -- & -- & -- \\
Ring Stacking & Rigid & 89 & -- & -- & -- \\
Color Block Sort. & Sem. & 97 & -- & -- & -- \\
Water Pouring & Fine & 27 & -- & -- & -- \\
Power Cord & Fine & 36 & -- & -- & -- \\
Obj.-to-Basket & Sem. & 100 & -- & -- & -- \\
Pencil Case & Fine & 26.5 & -- & -- & -- \\
\end{xltabular}
}

\subsection{Evaluation Task Descriptions and Scoring Rubrics}\label{sec:scoring_rubrics}

Real-robot evaluation in this report covers 31 tasks. All tasks follow a unified step-wise scoring protocol: each task has a maximum score of 10, accumulated step by step according to key manipulation stages. We define task progress $= \text{actual score} / \text{maximum score} \times 100$ as the fine-grained evaluation metric. Each task is evaluated over 10 trajectories; success rate is defined as the proportion of trajectories that fully complete all steps. Details can be found in table~\ref{tab:scoring_rubrics}.

\newpage
{\small
\begin{xltabular}{\textwidth}{@{}>{\raggedright\arraybackslash}p{0.19\textwidth}>{\raggedright\arraybackslash}X>{\raggedright\arraybackslash}X@{}}
\caption{Real-robot evaluation task descriptions and scoring rules.}
\label{tab:scoring_rubrics} \\
\toprule
Task & Description & Scoring Rule \\
\midrule
\endfirsthead
\toprule
Task & Description & Scoring Rule \\
\midrule
\endhead
\midrule \multicolumn{3}{r}{\emph{Continued on next page}} \\
\endfoot
\bottomrule
\endlastfoot
Block Sorting & Sort 5 colored blocks by color into matching trays & 2 pts per correct placement, 5 blocks total \\
Fruit Sorting & Grasp specified fruit per instruction and place into designated tray & Grasp (4), correct tray (5), retract (1) \\
Number Ordering & Arrange number cards in correct sequence based on a pattern & 1 pt per correct card (7 total), 3 bonus for all correct \\
Switch Pressing & Move the arm to the light switch and press it & Move to position (3), press switch (4), retract (3) \\
Ring Stacking & Place a ring object onto a vertical pole & Grasp ring (3), move to pole (3), stack (3), retract (1) \\
Cup Grasping & Place a cup onto a plate; must first push plate to center & Push plate (3), upright cup (2), pick up (2), place (2), retract (1) \\
Towel Folding & Sequentially pick up, spread, fold, and place 2 towels & Per towel: pick up (1), spread (1.5), fold (1.5), place (1) \\
Table Setting & Set a Western-style table: fold napkin, place plate, cutlery, etc. & Fold napkin (2), 1 pt per step (7 steps) \\
Paper Shredding & Take paper from rack and feed into shredder & Per sheet: grasp (1) + feed (2); 3 sheets, retract (1) \\
Charger Plugging & Pick up charger plug and insert into outlet; sub-cm alignment & Pick up (2), grip (3), insert (4), retract (1) \\
Flower Arranging & Pick up 3 flowers and place into a vase & Per flower: grasp (1.5) + place (1.5); 3 flowers, retract (1) \\
Package Sorting & Extract packages, flip barcode up, place in area & Per pkg: pick up, place, barcode up; 6 packages \\
Toy Basket Plcmt. & Pick up 4 toys and place into basket without tipping & 2 pts per toy (4), basket stable (1), retract (1) \\
Rope Tightening & Bimanual: grasp slack rope and pull taut & Each hand approaches (2), grasp rope (4), pull taut (2) \\
Bean Pouring & Pour beans from bag into basin; bimanual & Move basin (2), bag above (2), grasp bottom (3), pour (3) \\
Table Wiping & Pick up cloth and wipe grease stains & Approach (2), pick up (2), wipe all (5), return (1) \\
Pot Lid Covering & Pick up pot lid and place on pot & Approach (2), pick up lid (3), cover (4), return (1) \\
Cup Triangle Stack. & Invert 3 cups in triangle: 2 bottom, 1 top & Per cup: pick up (1) + position (1--3); top (3), retract (1) \\
Spoon-in-Bowl & Move bowl to center, place spoon in bowl & Bowl (2), move (2), spoon (2), place in bowl (3), retract (1) \\
Glasses Rack Plcmt. & Pick up glasses, adjust, place on rack & Pick up (2), center (2), adjust (2), rack (3), retract (1) \\
Drawer Organization & Place items into correct drawer levels by category & Per drawer: open (1) + place (1) + close (1); 3 drawers, retract (1) \\
Color Block Sort. & Place RGB blocks onto matching color patches & 3 pts per correct, 3 blocks, retract (1) \\
Water Pouring & Pour water from kettle into cup & Move cup (2), kettle (2), pour (3), set down (2), retract (1) \\
Power Cord Plug. & Bimanual: left picks cable, passes to right, inserts & Left grasp (3), right receive (3), insert (3), retract (1) \\
Obj.-to-Basket & Pick up 4 objects and place into basket & Per object: pick up (1) + place (1.25); 4 objects, retract (1) \\
Pencil Case Pack. & Bimanual: open zipper, insert items, close & Open (3.5), 1 pt per item (3), zip (1.5), retract (1) \\
Earphone Sorting & Find earphones in clutter, place in basket & Identify (2), pick up (3), place (4), retract (1) \\
Shape Sorting & Sort 3 objects by shape to positions & 3 pts per correct, 3 objects, retract (1) \\
Seq.\ Button Press & Press 3 buttons in instructed order & 3 pts per correct press, 3 buttons, retract (1) \\
Object Matching & Match object pairs, place at positions & Per pair: grasp (2) + place (2.5); 2 pairs, retract (1) \\
Fruit Basket Plcmt. & Place specified fruits into basket & 3 pts per correct fruit, 3 fruits, retract (1) \\
\end{xltabular}
}